# Uncertainty Quantification of Wind Gust Predictions in the Northeast United States: An Evidential Neural Network and Explainable Artificial Intelligence Approach

Israt Jahan[1], John S. Schreck[4], David John Gagne[3,4], Charlie Becker[4], Marina Astitha[1,2,3]

[1]*School of Civil and Environmental Engineering, University of Connecticut, Storrs, CT, USA*

[2]*Eversource Energy Center, University of Connecticut, Storrs, CT, USA*

[3]*NSF National Center for Atmospheric Research (NCAR), Research Applications Laboratory, Boulder, CO, USA*

[4]*NSF National Center for Atmospheric Research (NCAR), Computational & Information Systems Lab, Boulder, CO, USA*

*Corresponding author*: Marina Astitha, marina.astitha@uconn.edu

ABSTRACT

Machine learning algorithms have shown promise in reducing bias in wind gust predictions, while still underpredicting high gusts. Uncertainty quantification (UQ) supports this issue by identifying when predictions are reliable or need cautious interpretation. Using data from 61 extratropical storms in the Northeastern USA, we introduce evidential neural network (ENN) as a novel approach for UQ in gust predictions, leveraging atmospheric variables from the Weather Research and Forecasting (WRF) model. Explainable AI techniques suggested that key predictive features contributed to higher uncertainty, which correlated strongly with storm intensity and spatial gust gradients. Compared to WRF, ENN demonstrated a 47% reduction in RMSE and allowed the construction of gust prediction intervals without an ensemble, successfully capturing at least 95% of observed gusts at 179 out of 266 stations. From an operational perspective, providing gust forecasts with quantified uncertainty enhances stakeholders' confidence in risk assessment and response planning for extreme gust events.

## 1. Introduction

Wind gusts arise from the transfer of high-momentum air to the surface (Kahl, 2020) and are influenced by various small-scale phenomena, including wind shear, turbulence, surface roughness, topography, and surface heating by solar radiation. These complex interactions pose a significant challenge for gust prediction, as the turbulent eddies producing gusts typically fall below the spatial resolution that standard observation networks and mesoscale numerical models can capture (Fovell and Cao, 2014). The impacts of wind gusts can be multifaceted; ranging from uprooting trees and toppling electric poles, which can cause power outages, to creating unsafe driving conditions during high wind events. Therefore, accurate gust prediction is crucial for mitigating risks to infrastructure, transportation, and public safety, especially in regions prone to severe weather. However, even the state-of-the-art numerical weather prediction (NWP) models are susceptible to uncertainty in their predictions, which can arise from various sources such as model limitations, initial and boundary data quality, parameterization of sub-grid scale processes or inherent variability of the atmosphere. Understanding and quantifying this uncertainty is critical for decision-makers to assess gust forecast reliability and make informed, risk-based decisions such as trimming trees near power lines or deploying crews to vulnerable areas in advance.

Traditionally, uncertainty in weather prediction has been addressed through ensemble forecasting of physics-based NWP models which generate probabilistic forecasts using ensembles of deterministic forecasts by perturbing the initial and boundary conditions or using different model configurations, providing a range of possible outcomes (Leith, 1974). While this method leverages the true physics of atmospheric/oceanic motion, the deterministic numerical model ensembles come with considerable computational costs and often lack proper uncertainty calibration (Vannitsem et al., 2018) arising from uncertainty in the initial conditions or model parameterizations. Another approach to address uncertainty quantification (UQ) is the use of statistical methods such as ensemble model output statistics (EMOS) (Gneiting et al., 2007) and Bayesian model averaging (BMA) (Raftery et al., 2005). EMOS assumes a parametric distribution and adjusts the parameters of the distribution (mean and variance) using statistical regression techniques. BMA accounts for uncertainty by assigning weights to each ensemble member that reflects its likelihood or performance based on past observations. While widely used, these statistical approaches can be computationally expensive, requiring multiple runs of NWP models and may not scale well to complex, high-dimensional problems such as wind gust prediction.

Using machine learning (ML) to address prediction uncertainty in atmospheric science has gained much traction in recent years (Haynes et al., 2023; McGovern et al., 2017), especially within the framework of weather forecast post-processing (Haupt et al., 2021; Schulz and Lerch, 2022; Mohammadi et al., 2023). Primo et al. (2024) compared the performance of model output statistics (MOS) and neural network (NN) based approaches in postprocessing wind gusts and concluded that NN approaches showed better calibration and higher accuracy of gust forecasts compared to traditional MOS methods. Chen et al. (2024) used a generative ML approach for multivariate postprocessing of ensemble weather forecasts that outperformed traditional two-step methods. Unlike the two-step process, which separately calibrates marginal distributions and then restores multivariate dependencies, the generative ML method simultaneously corrects both without relying on parametric assumptions. Moosavi et al. (2021) used Random Forest (RF) and NN to predict and reduce uncertainty in the WRF model precipitation forecasts resulting from the interaction of several physical processes included in the model.

While uncertainty estimates are essential for any predictive model, breaking down uncertainty into its aleatoric and epistemic components can help determine the mitigation strategies to reduce uncertainty. This differentiation can help prioritize whether efforts should focus on improving input data quality or refining the model structure. It is worth noting that definitions of aleatoric and epistemic uncertainty may vary slightly across disciplines, such as mathematics versus ML (Bevan, 2022; Hüllermeier and Waegeman, 2021). Since this paper utilizes ML to quantify uncertainty, we follow the definitions commonly used in the ML literature (Haynes et al., 2023; Schreck et al., 2024). Aleatoric uncertainty refers to the inherent data randomness, including internal variability of physical processes, observation error or physical model error. For example, we may have different values for wind gusts (target) given the same surface wind speed or terrain height (features) due to internal variability of physical processes. Epistemic uncertainty can arise from incorrect ML model structure, poor model parameter estimates and insufficient training data (Haynes et al., 2023; Kendall and Gal, 2017; Schreck et al., 2024). High aleatoric uncertainty suggests a weak relationship between the features and the target and can only be reduced by incorporating more informative features (Herman and Schumacher, 2018). High epistemic uncertainty can be reduced by adding more data in underrepresented regions of the input space or through exploration of different model architectures and hyperparameter configurations (Schreck et al., 2024). While hyperparameter optimization can be beneficial, reducing epistemic uncertainty remains challenging with sparse training data or the model encountering out-of-distribution inputs. In recent years, evidential neural networks (ENNs) have been successfully applied to UQ (Akihito Nagahama, 2023; Amini et al., 2020; Gao et al., 2024; Sensoy et al., 2018; Soleimany et al., 2021; Ulmer et al., 2023) for a wide range of tasks. ENN offers an efficient way to estimate both aleatoric and epistemic uncertainty at a reasonable computational cost, since it uses a single deterministic NN that reformulates the prediction task to estimate parameters of a higher-order evidential distribution, based on principles from Bayesian data analysis (Gelman et al., 2014). ENN provides both point predictions and uncertainty estimates in a single forward pass, without the need for computationally expensive ensembles. Additionally, by explicitly decomposing total uncertainty into aleatoric (input data-related) and epistemic (model-related) components, ENN facilitates better understanding of the source of the uncertainty. The mathematical formulations detailing how ENN calculates these uncertainties are provided in the Supplement (Section 1).

In a recent study by Jahan et al. (2024) (hereafter referred to as J24), wind gust predictions from the WRF Unified Post Processor (UPP) were found to be misaligned with observations from 61 low-pressure storms analyzed in the Northeastern (NE) US. ML models like RF and XGBoost, along with generalized linear models, outperformed WRF-UPP, but still underpredicted high gusts, especially those exceeding 25 m/s. The problem of underprediction of high gusts, primarily due to the right-skewed gust distribution (Fig. 1c), still persisted even after increasing the dataset from 48 to 61 storms. Additionally, the learning curves for the XGB model-the best performer- showed that after incorporating 30 storms, the changes in the error metrics with further addition of storms were no longer statistically distinguishable. Therefore, we kept the number of storms to 61 in this study, though some older storms from J24 were replaced due to the unavailable gust records from the New York State Mesonet (NYSM), which is a new addition in this study.

In this work, we introduce evidential deep learning as a novel UQ approach for wind gust prediction. While NWP models like WRF often overestimate gust magnitudes and lack forecast confidence, most existing post-processing techniques emphasize primarily on bias correction or deterministic accuracy, without offering a structured way to quantify uncertainty. This study aims to fill that gap by evaluating the potential of ENN to simultaneously correct WRF gust overpredictions and provide a more nuanced understanding of prediction confidence. To the best of our knowledge, no prior study has focused on decomposing wind gust prediction uncertainty into aleatoric and epistemic components. Our structured UQ approach can better support decision-making in weather-sensitive sectors such as energy, transportation, and emergency response. In addition, we have employed explainable AI (XAI) techniques to understand and trust the results and decision-making processes of ENN in a post-prediction setting and identify challenges and limitations for gust and gust uncertainty predictions (Flora et al., 2024; Mamalakis et al., 2022, 2023; McGovern et al., 2019). The paper is structured as follows. Section 2 describes the types of data: gust observations used as targets and the WRF variables used as predictors. Section 3 explains the methodology and Section 4 contains the discussion of the results. The conclusions follow in Section 5.

## 2. Data

We used two types of data in this study: observational wind gusts (hourly) and WRF model output (hourly). Similar to J24, we used 61 extratropical storms spanning between 2017 to 2021 that demonstrated a wide range of observed gusts (Fig. 1c). All the storms were low-pressure systems accompanied by cold fronts, and some of them also had warm, stationary, and occluded fronts co-occurring with the cold fronts. The list of storms is provided in Table A1 (Appendix A).

### *2.1. Observations*

Wind gust observations originated from two databases: the Global Hourly Integrated Surface Database (ISD) of NOAA and the NYSM network (Brotzge et al., 2020). ISD consists of observations from a variety of systems such as the automated surface observing system (ASOS), automated weather observing system (AWOS), METAR, coastal marine automated network (C-MAN), buoys, and several others (Smith et al., 2011). Most of the stations within our study region are ASOS and AWOS, along with a limited number of observatories, C-MAN stations, and buoys. Even though observations are compiled from various sources to create ISD, all observations go through a series of validity checks, extreme value checks, internal consistency checks, and temporal (check another observation for the same station) continuity checks (Lott, 2004). Gust observations from ISD are instantaneous values at each hour.

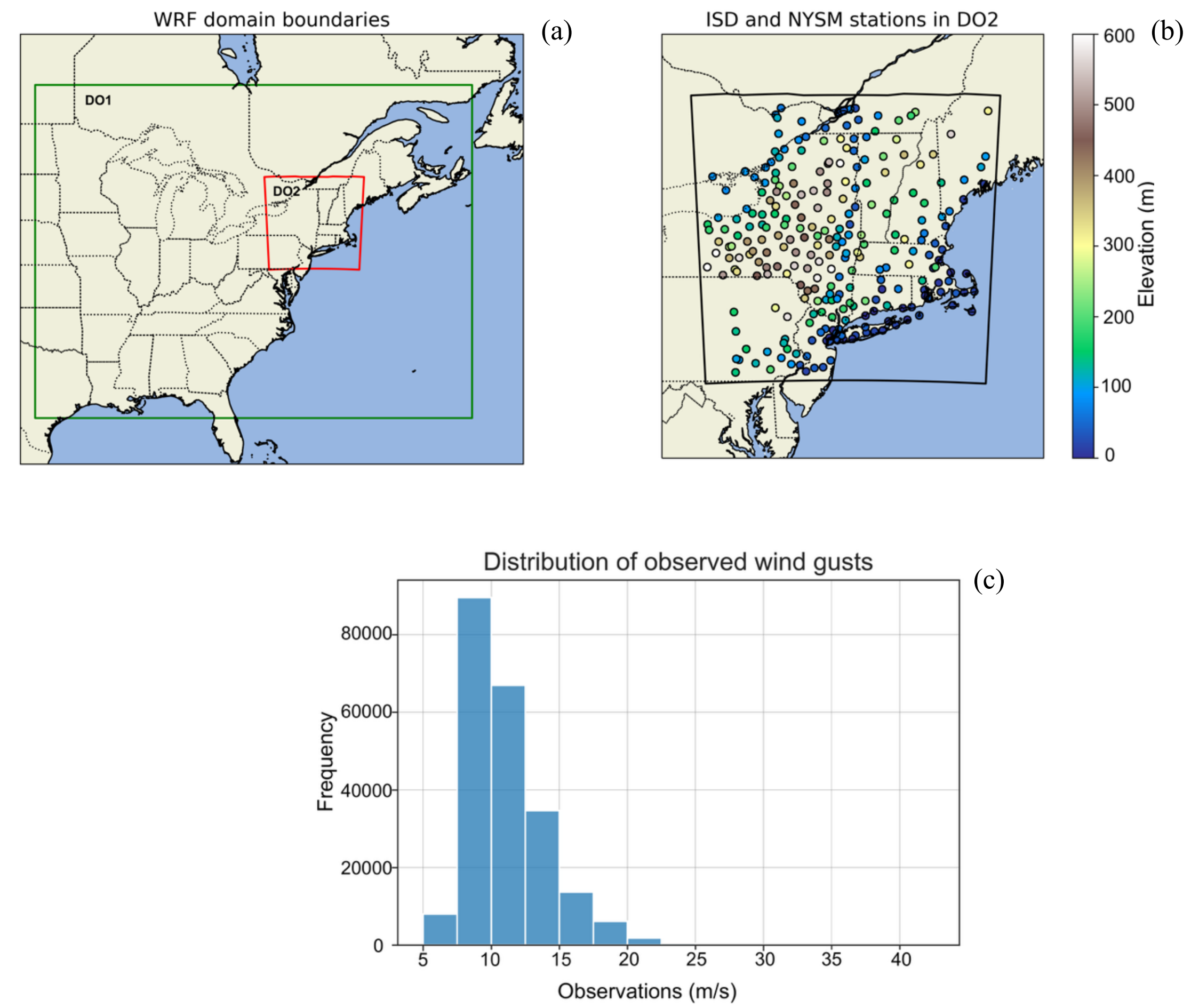

**Fig. 1.** (a) WRF model domains with 12 km grid spacing (DO1) and 4 km grid spacing (DO2). (b) The inner domain (DO2) with available ISD and NYSM weather stations. The color bar represents the elevation of the weather stations. (c) Frequency distribution of observed gusts for 61 storms used in this study.

The NYSM is an advanced weather station network consisting of 126 standard surface weather stations distributed across the New York state with an average spacing of 27 km (Brotzge et al., 2020). Integrating gusts from NYSM alongside ISD allowed us to leverage the higher density of stations within our region of interest, which is crucial for gridded gust predictions. Sparse station coverage can limit the ability of the model to capture localized variations in wind gusts, as the model primarily learns from data available at discrete weather stations during training. NYSM propeller gusts are available as 5-minute averages on the NYSM website (https://nysmesonet.org/weather/requestdata). To derive the hourly gust values (e.g., gust at 05 UTC), three specific time points were considered: 10 minutes before the hour (e.g., 04:50 UTC), 5 minutes before the hour (e.g., 04:55 UTC), and at the top of the hour (e.g., 05:00 UTC). The maximum gust value among these three measurements was selected as the gust for that hour. Similar to ISD, the NYSM data is passed through automated quality assurance (QA) and quality control routines which include a variety of filters and tests, such as range filters, step tests, and similarity tests. In addition, manual QA is performed including daily, monthly, and annual reviews by the QA manager. ISD and NYSM stations within 4 grid cells (16 km) of the lateral boundaries of the inner domain (DO2) were excluded so that boundary conditions would not affect our analysis. It is to be noted that wind gusts occur intermittently and as such, gust observations were available only for those hours during a storm when gusts were recorded at a given station. Also, not all stations reported gusts at the same hour during a storm, since storm systems passed across the stations at different hours while moving over the study region. Additional details on pairing of observed gusts with WRF features are provided in Section 3.2.

### *2.2. WRF model data*

The storms were simulated using WRF v4.2.2 which extended for a total duration of sixty hours for each storm, including a twelve-hour spin-up time. For each storm, model output within the spin-up period was discarded from the analysis resulting in hourly outputs for 48 hours. Global Forecast System (GFS) analysis data with a spatial resolution of 0.5 degrees was used to provide the initial and boundary conditions for the WRF model. The model configuration included the same parameterization schemes used by Jahan et al. (2024): Thompson et al. (2008) scheme for microphysics; the Rapid Radiative Transfer Model (Mlawer et al., 1997) for longwave radiation; the Goddard shortwave scheme (Chou and Suarez, 1994) for shortwave radiation; the YSU scheme (Hong et al., 2006) for the planetary boundary layer; the Noah land surface model; and the Revised MM5 surface layer scheme.

The model incorporated 51 vertical levels and two nested domains, utilizing a two-way nesting approach. While the outer domain with 12 km grid spacing covered a large part of the US, the inner domain was set up at 4 km grid spacing focusing the NE US (Fig. 1a). As our selected storms passed over the NE US, we used meteorological variables from the hourly output of the inner model domain (D02). Table 1 lists the meteorological variables extracted from WRF and used as features for ENN.

**Table 1. WRF-derived atmospheric variables used as input features for training ENN.**

| Variable | Description | Units |
|---|---|---|
| WS_10m | Wind speed at 10 m height | m/s |
| WS_850mb | Wind speed at 850 mb pressure level | m/s |
| WS_950mb | Wind speed at 950 mb pressure level | m/s |
| PBLH | Planetary boundary layer height | km |
| Ustar | Friction velocity | m/s |
| WindDC (sin) | Sine of surface wind direction | - |
| WindDC (cos) | Cosine of surface wind direction | - |
| Terrain_height | Terrain height | m |
| Lapse_rate (sfc_1km) | Lapse rate (surface to 1 km height) | °C/km |
| Lapse_rate (sfc_2km) | Lapse rate (surface to 2 km height) | °C/km |
| yday | Cosine-transformed day of the year: $\left\{2\pi[\frac{t-1}{365}]\right\}$ where t is day of the year (Schulz and Lerch 2022) | - |

The feature selection process was similar to J24. WRF-derived weather variables that are physically relevant to wind gust phenomena were selected as features. Many of the features, listed in Table 1, are consistent with previous studies (Coburn et al., 2022; Coburn and Pryor, 2022; Schulz and Lerch, 2022). Our initial feature set also included variables such as 2-meter temperature, 2-meter potential temperature, surface pressure, and potential temperature lapse rates at various levels (surface to 1 km, surface to 2 km, and surface to planetary boundary layer height). We removed 2-meter temperature due to its near-perfect correlation with 2-meter potential temperature, and also excluded potential temperature lapse rates due to strong negative correlations with temperature lapse rates. Permutation importance analyses across multiple trials further indicated that both 2-meter potential temperature and surface pressure had consistently low importance, leading us to exclude them. While we used wind speed at different vertical levels as features, we limited wind direction to the surface level. Our initial features included only surface wind speed and direction, however, consistent high importance of surface wind speed in feature importance analysis led us to include wind speed at 850 hPa and 950 hPa which are also used in gust estimation in ERA5 reanalysis (ECMWF, 2024). In contrast, surface wind direction ranked as one of the least important predictors and therefore, we did not consider wind direction at different vertical levels. Our final selected WRF variables closely align with those in J24, with a few modifications. In this study, we experimented with two additional features: forecast hour and cosine-transformed day of the year. Permutation feature importance (PFI) and partial dependence plots (PDPs) revealed that forecast hour had little to no impact on predicted gusts and estimated uncertainty, while the cosine-transformed days of the year had a positive effect on both. Therefore, we added cosine-transformed days of the year as a feature in this study. Fig. S8 (Supplement) shows the correlation matrix among the WRF feature variables and the target wind gust.

## 3. Methodology

To train ENN, we used the Generalized Uncertainty for Earth System Science (GUESS) python package developed by the Machine Integration and Learning for Earth Systems (MILES) group at NSF NCAR. MILES-GUESS (https://github.com/ai2es/miles-guess/tree/main/mlguess) supports both Keras and PyTorch to compute uncertainty measures. All analyses is this study were conducted using Keras. ENN, which is designed following a multi-layer perceptron model structure, enables the estimation of the aleatoric and epistemic component of total uncertainty by applying the law of total variance. A brief overview of evidential regression and the law of total variance are provided in the Supplementary Documentation (Section 1).

### *3.1. Evaluation metrics*

We used a variety of statistical metrics to assess the model performance, categorized into two groups. The first group of metrics focused on assessing the error in predicted gusts, such as mean bias, mean absolute error (MAE), root mean squared error (RMSE), centered root mean square error (CRMSE), and Pearson correlation coefficient (Table S1 in Section 2.1 of the Supplement). The second group of metrics were used to assess the calibration of uncertainty estimates and for that purpose, we adopted the metrics and graphics outlined by Haynes et al. (2023) and Schreck et al. (2024), such as the discard fraction diagram, spread-skill relationship and probability integral transform deviation (PITD) skill score. Definitions of these three metrics are provided in Section 2.2 of the Supplement.

In addition, we used prediction interval coverage probability (PICP) to evaluate the quality of uncertainty estimates against observations. PICP helps assess uncertainty estimates by constructing prediction intervals at a specified confidence level and calculating the fraction of observations that fall within these intervals (Sluijterman et al., 2024). The approach is outlined below:

I. First, the prediction intervals (PIs) at a desired confidence level (e.g., 95%) were constructed using the mean prediction ($\mu$) and total uncertainty estimate ($\sigma$) for each hourly predicted gust following Eq. 1-3 (Coskun, 2024; Ramachandran and Tsokos, 2020).

$$PI = \mu \pm z\sigma \qquad \text{(Eq. 1)}$$

$$L_{PI} = \mu - z\sigma \qquad \text{(Eq. 2)}$$

$$U_{PI} = \mu + z\sigma \qquad \text{(Eq. 3)}$$

Here, $\mu$ is the mean gust prediction from ENN, $\sigma$ is the total uncertainty in the prediction from ENN, $z$ is the z-score taken from statistical tables corresponding to the desired confidence level (e.g., z=1.96 for 95% confidence level), $L_{PI}$ and $U_{PI}$ are the lower and upper boundaries of the PIs respectively.

II. After constructing PIs for each hourly predicted gust, PICP was computed following Eq. 4.

$$PICP = \frac{1}{n}\sum_{i=1}^{n} C_i \qquad \text{(Eq. 4)}$$

Here, $n$ is the number of observations and $C_i$ is a boolean variable with a value of 1 when the observed gust falls within the range $[L_{PI}, U_{PI}]$ or 0 when it is outside the range. Thus, PICP denotes the fraction of observations that fall within the PIs computed at a certain confidence level. For example, PICP of 0.75 at 95% confidence level means that 75% of the observations are within the estimated PIs, which are $\mu \pm 1.96\sigma$, at 95% confidence level. The higher the confidence level, the wider the PIs are.

### ***3.2. Model training and validation***

Our selected 61 storms occurred between 2017 to 2021 with the caveat that the number of storms in each year was uneven. The total number of gust observations for the 61 storms was 221,359. We split the data into training (60%), validation (20%), and test (20%) sets, ensuring that no single storm was divided across multiple sets. The storms were ordered chronologically, with the earliest 42 storms (131,889 observations) used for training, roughly corresponding to 60% of the total data. The next 8 storms (44,585 observations) were used for validation, and the latest 11 storms (44,885 observations) served as the test set, roughly representing 40% (20% validation and 20% test) of the total data (Table A1).

To form the final training and validation datasets, the WRF variables (Table 1) were bilinearly interpolated to the weather station locations (Fig. 1b) based on their latitude and longitude. For each station, the value of a variable was estimated as a weighted average of the four nearest WRF grid points based on the station's relative distance to those surrounding points. These interpolated variables were then paired with observed gusts based on the matching time, aligning WRF output time with observed gust time. For model inference, we used gridded WRF variables for the test storms and made hourly predictions at the WRF grid cells excluding the ocean part over the domain due to not having any stations there in the observation dataset. Since gridded gust observations were unavailable, we evaluated the model by interpolating the gridded predictions to the station locations with observed gusts.

Hyperparameter tuning was performed using the Earth Computing Hyperparameter Optimization (ECHO) tool (Schreck and Gagne, 2021), a hyperparameter optimization package built with Optuna. Since our goal was to optimize the model for both predicted gusts and calibrated uncertainty estimates, we used the multi-objective optimization version of the Tree-structured Parzen Estimator (TPE) (Bergstra et al., 2011), called the MOTPE algorithm (details provided in the Supplement, Section 3). After 500 trials, the results from the hyperparameter tuning converged. The hyperparameters explored for the evidential model and their optimal values are provided in Table S2.

After hyperparameter tuning, we combined the 42 training storms and 8 validation storms into a single dataset of 50 storms. We then performed five-fold cross-validation with a fixed validation size of 10 storms. In each iteration, the model was trained on 40 storms and validated on the remaining 10, with the validation storms rotating across folds. Each storm was used exactly once for validation and included in training during the other iterations. This approach allowed us to evaluate model performance across different subsets of storms. We observed similar performance in different validation folds, suggesting that the model was not overfitted. Finally, the model was retrained on all 50 storms using the optimal hyperparameters selected via ECHO and applied to the held-out test storms. The methodology is shown in Fig. 2.

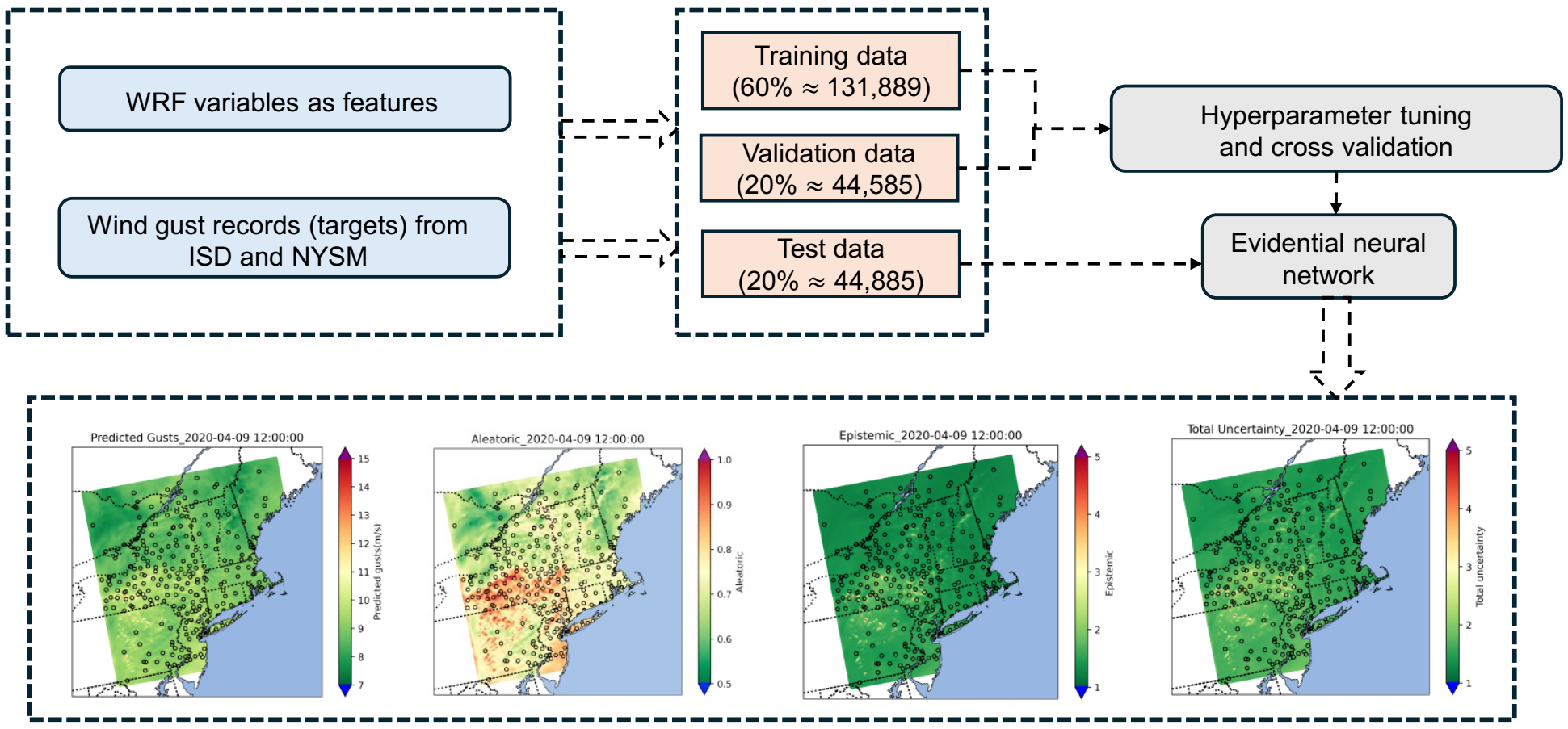

**Fig. 2.** Illustration of the WRF-ENN workflow.

The model showed different degrees of calibration for the three types of uncertainties (Fig. S1 in the Supplement). The spread-skill diagram showed a stronger 1-1 relationship for epistemic and total uncertainty, suggesting better calibration of the model in terms of epistemic and total uncertainty compared to aleatoric uncertainty (Fig. S1b). Lack of model calibration according to aleatoric uncertainty was also observed in the discard fraction plot (Fig. S1a) and PIT histogram (Fig. S1c), a similar pattern noted by Schreck et al. (2024). Fig. S1b also showed that epistemic uncertainty was approximately three times higher than aleatoric uncertainty, making it the dominant contributor to total uncertainty in predictions. Therefore, for the sake of brevity, our analysis focused on total uncertainty.

## 4. Results and Discussion

To assess the performance of the ENN against the physics-based model, we used post-processed gusts from WRF-UPP. Fig. 3a and 3b show the wind gust predictions from ENN and WRF-UPP respectively, against observations for 11 test storms. While WRF-UPP shows substantially high overprediction, ENN shows underprediction of relatively high gusts (Fig. 3). J24 used RF, XGBoost and Generalized Linear models to predict wind gusts where similar underprediction of high gusts was observed. This is a common limitation of ML models in capturing extreme values due to imbalance between extreme and typical values in a training dataset. Nevertheless, despite ENN's tendency to underpredict high gusts, the gust distributions between observations and ENN's predictions are still much closer at each hour of the storm compared to WRF-UPP as shown in Fig. 4.

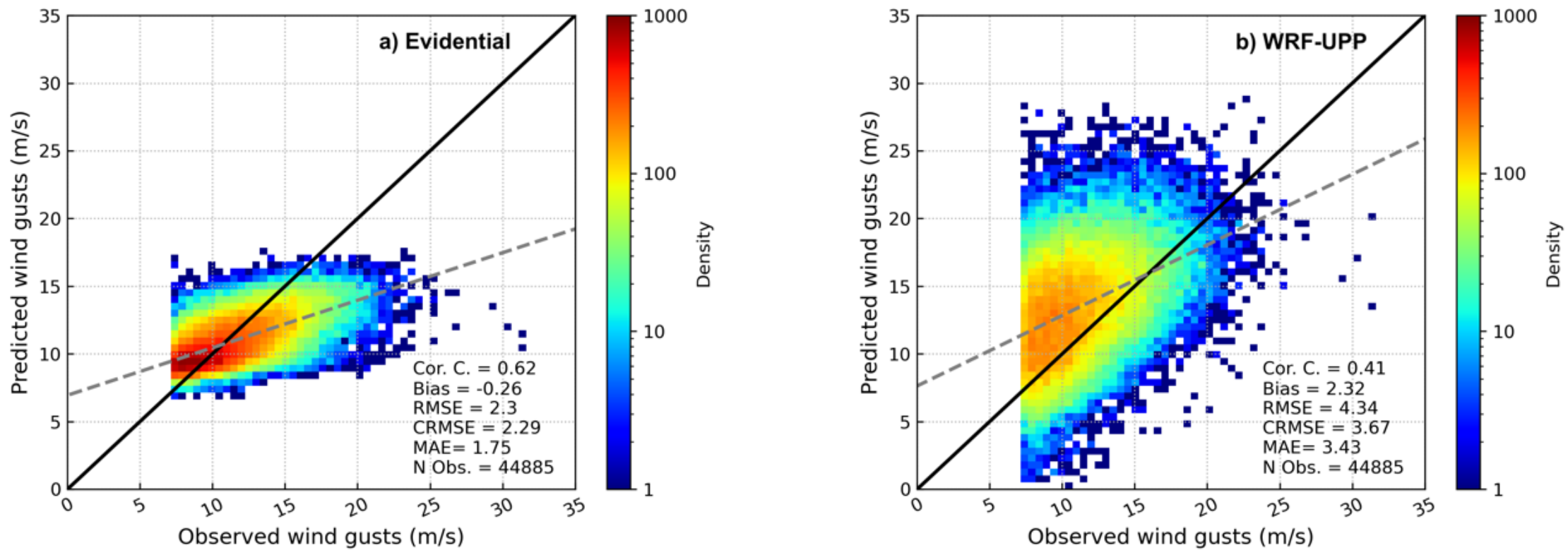

**Fig. 3.** Wind gust predictions by (a) ENN, and (b) WRF-UPP. Bias, RMSE, CRMSE and MAE have the same units as wind gust, which is m/s.

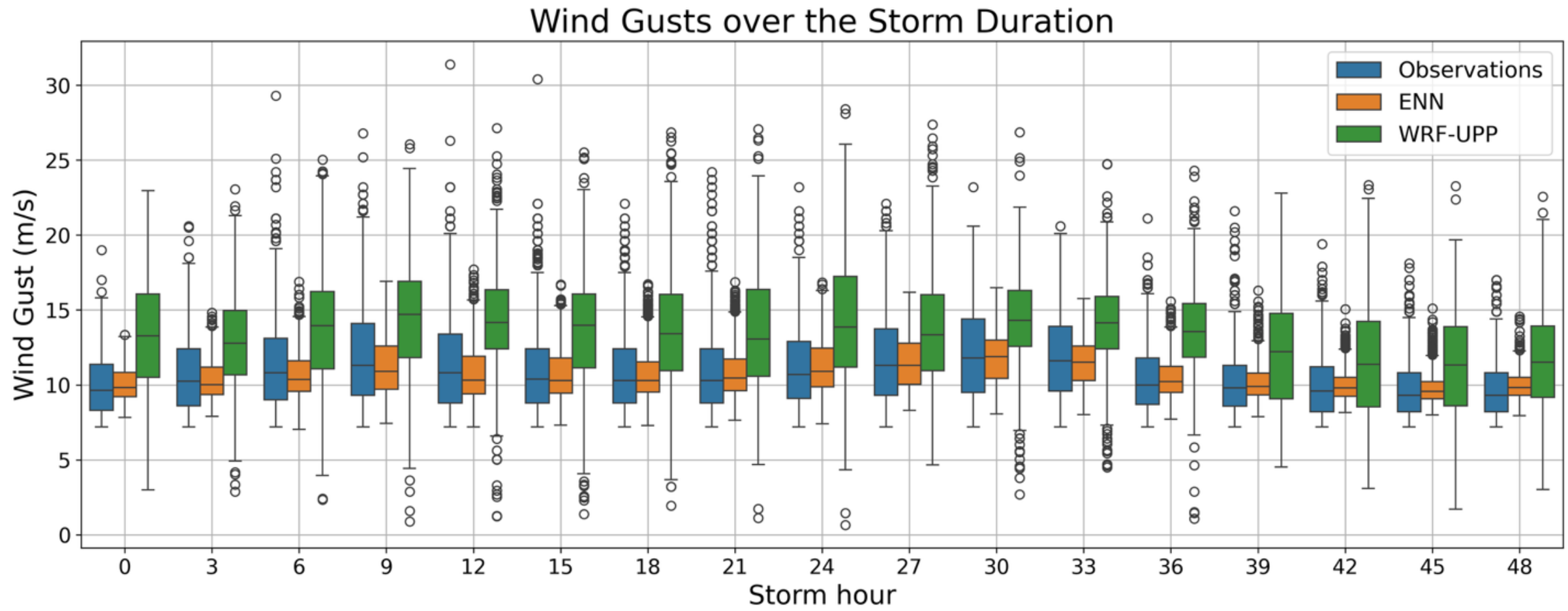

**Fig. 4.** Box plots of wind gusts at 3-hour intervals for the test storms. Hour 0 indicates the storm onset and hour 48 marks the end of the 48-hour storm period. Each box comprises gusts at all stations within the study area for the 11 test storms.

To investigate if data augmentation can mitigate the problem of underprediction of high gusts, we attempted synthetic minority over-sampling technique (SMOTE) (Chawla et al., 2002). We used the library SMOGN (https://github.com/nickkunz/smogn) which is a python implementation of synthetic minority over-sampling technique for regression with gaussian noise based on Branco et al. (2017). It did not improve the prediction of the upper tail, in fact, the overall error statistics were slightly better for ENN without any data augmentation. We also conducted error analysis stratified by gust intensity percentiles and observed no improvement in the higher gust percentiles. Details are provided in Section 4 of the Supplementary material (Fig. S2-S5 and Table S3). Since extreme gusts are poorly represented with or without data augmentation, the model's epistemic uncertainty, arising from a lack of knowledge or data, tends to be higher for these cases (discussed further in Section 4.2) Given that it is quite unlikely to get a balanced gust dataset, UQ from ENN along with predictions, can facilitate decision making by providing insights about uncertain predictions.

### *4.1. Evaluation of Wind Gust Prediction Intervals (PIs) Constructed from Uncertainty Estimates*

While the calibration plots indicate that the model correctly sorted the data (Fig. S1a in the Supplement) and that there was a good correlation between predicted uncertainties (epistemic and total) and calculated RMSE (Fig. S1b in the Supplement), the predicted uncertainty values should still be interpreted with caution. Ovadia et al. (2019) compared various probabilistic deep learning methods and found that even slight distributional shifts in test data from the original training set can degrade the quality of uncertainty estimates. Schreck et al. (2024) also reported instances of inflated epistemic uncertainties that did not align with the range of their target variable. Consistent with previous findings on ENN, we also observed occasional uncertainty estimates that exceeded the expected bounds of wind gusts. Therefore, before computing the prediction intervals (PIs) following the approach described in Section 3.1, we performed a percentile distribution analysis of total uncertainty to determine the proportion of uncertainty estimates that substantially deviated from the target variable's range (Fig. 5). The entire distribution of total uncertainty estimates over all test storms can be found in the Supplement (Fig. S10).

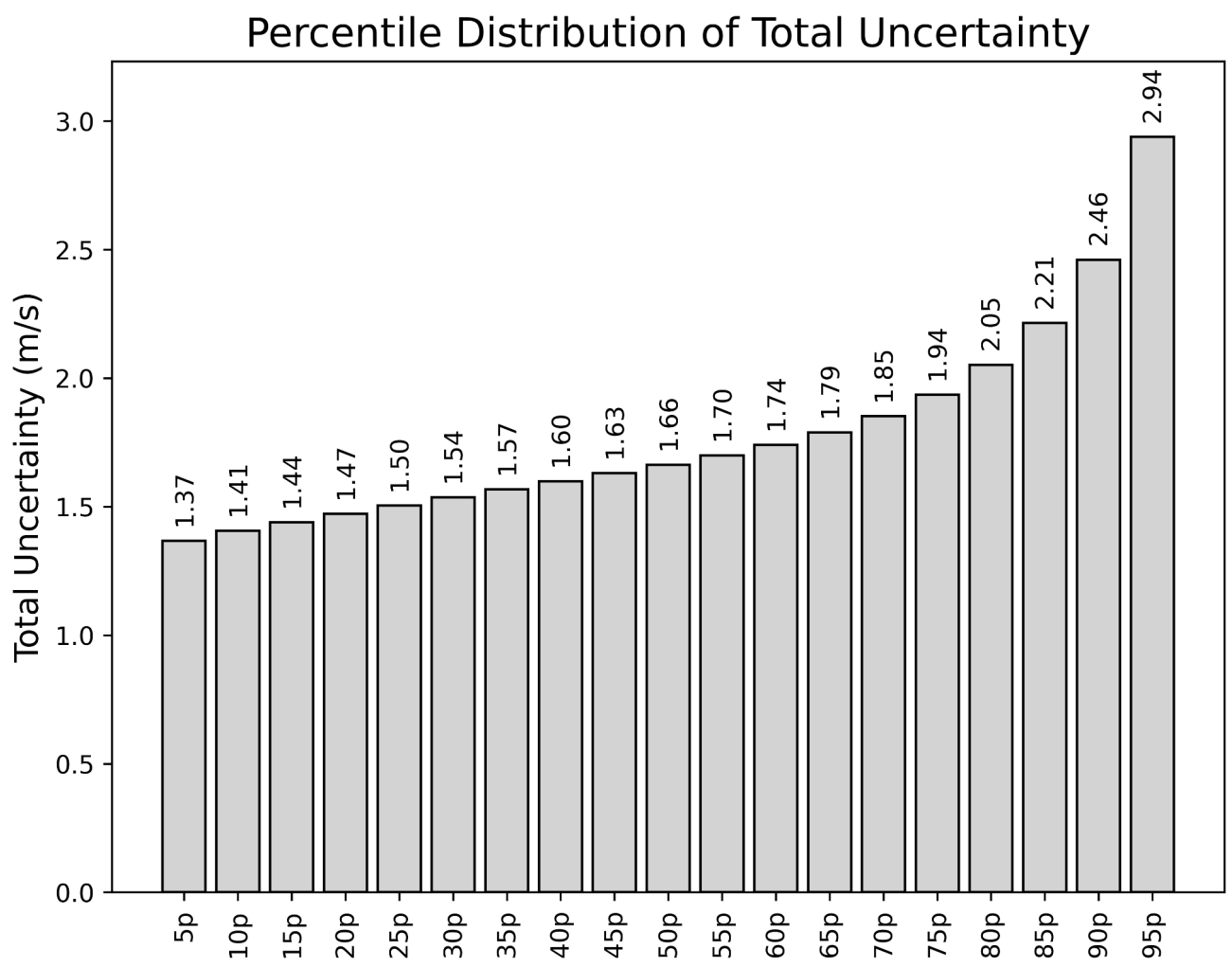

**Fig. 5.** Percentile distribution of total uncertainty in predicted gusts for the 11 test storms.

Fig. 5 shows that the total uncertainty (σ) in predicted wind gusts ranged from 1.37 m/s at the 5$^{th}$ percentile to 2.94 m/s at the 95$^{th}$ percentile. We designated any value surpassing the 95$^{th}$ percentile of total uncertainty as a 'highly uncertain prediction'. They were excluded from the PICP calculation to prevent the PICP value from being biased by large PIs that encompass the observations.

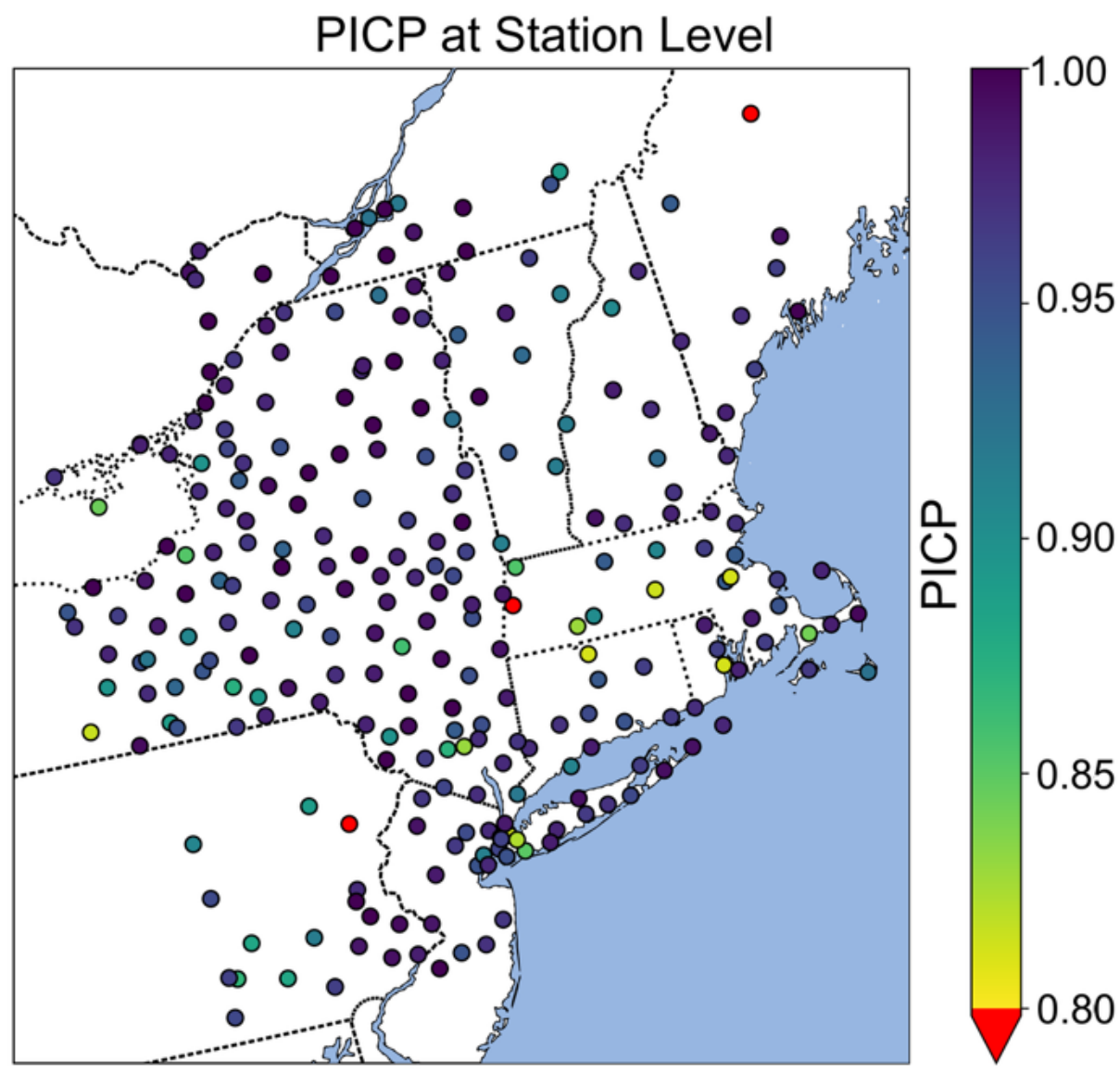

**Fig. 6.** PICP calculated for each station over the test storms at 95% confidence level.

We computed PICP by generating PIs at the 95% confidence level for each station following Eq. 1-4, repeating the process for all 266 stations (Fig. 6). It is important to clarify that the PICP calculation is not intended to compare the prediction accuracy among stations. As intermittent gusts result in different numbers of observations across stations, direct comparison of prediction accuracy among the stations is not appropriate. The goal was to assess whether the PIs derived from the total uncertainty estimates of ENN captured the actual observations. Our analysis showed that out of the 266 stations, 179 stations (approximately 67% of the total number of stations) had PICP values of 0.95 or higher. This suggests that, for the majority of the stations, the model's uncertainty estimates effectively captured the observed wind gusts. A note here that PICP greater than 0.95 was observed for stations that had relatively large number of gust records as well as stations having the least number of observed gusts.

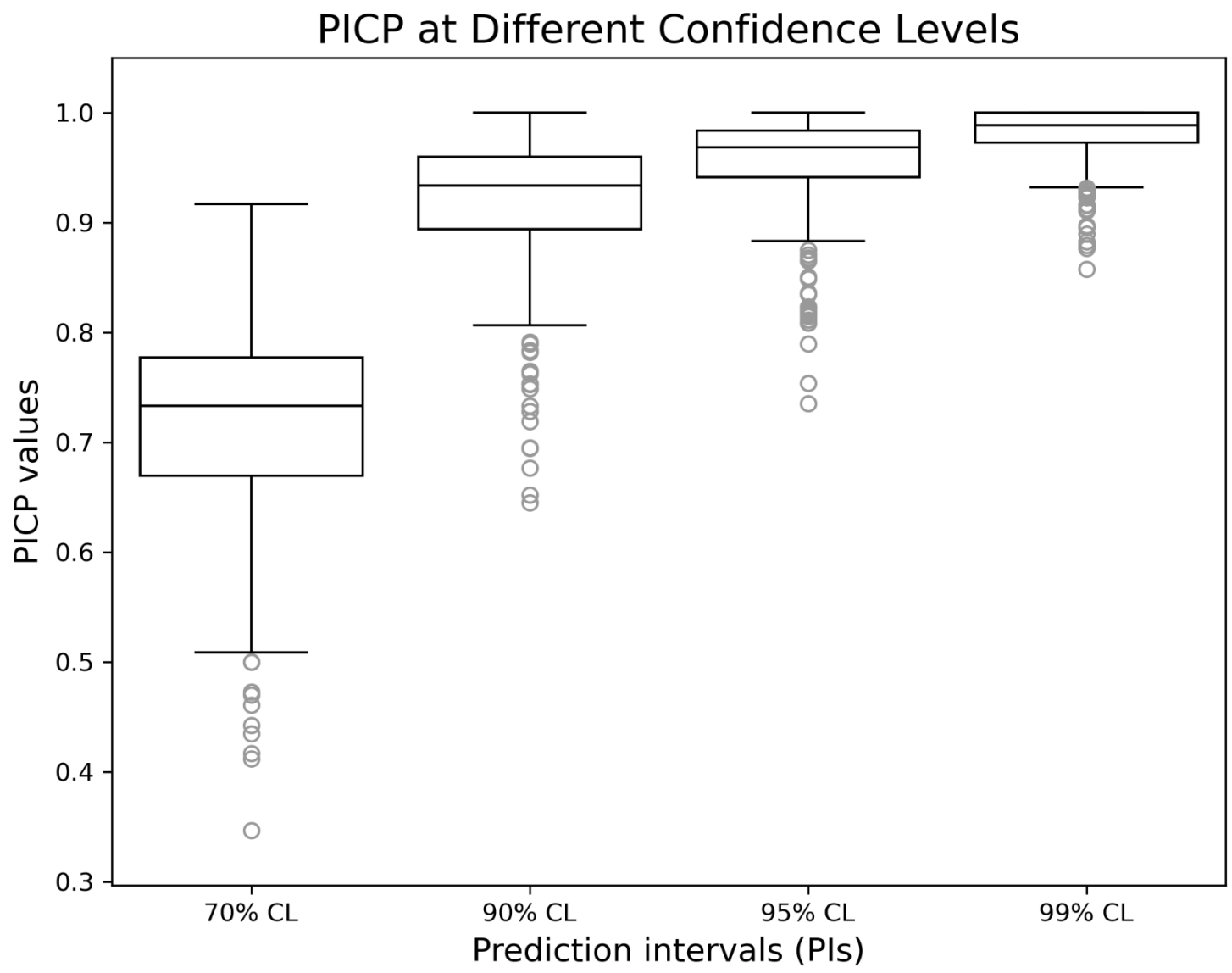

**Fig. 7.** PICP values over 266 stations at different confidence levels computed using Eq. 1-4. CL= confidence level. The z score in Eq. 1 corresponds to 1.04 for 70% CL, 1.65 for 90% CL, 1.96 for 95% CL and 2.58 for 99% CL. Each data point in the box plots corresponds to the PICP value of an individual station. For each station, PICP was determined using gust data from all test storms.

Fig. 7 illustrates the distribution of PICP values for PIs at 70%, 90%, 95% and 99% confidence levels. While 90%, 95% and 99% are commonly used thresholds in many studies, we also explored the variability in PICP at 70% confidence level, since it approximately corresponds to unit standard deviation (or total uncertainty) within the mean. At 70% confidence level, the PICP values across stations (excluding outliers) ranged from 0.51 to 0.92, indicating that the prediction intervals included the observed gusts in 51% to 92% of instances at these stations. As expected, higher confidence levels yielded increased PICP values across stations. For example, at 95% confidence level, PICP values ranged between 0.88 and 1, suggesting that at each station (excluding the outliers), 88% to 100% of the observations were captured by the intervals. These findings have practical implications for gust forecasting. From an operational perspective, issuing gust forecasts by defining the uppermost and lowermost probable boundaries of the forecast values solely by adding or subtracting the total uncertainty (as done for 70% CL) could lead to overconfidence in predictions. Such overconfidence might cause stakeholders to underestimate the risk of deviations, potentially leading to inadequate preparation or response to extreme gust events. Instead, a safer and more reliable approach would be to use a factored total uncertainty (as done for 90%, 95%, and 99% CLs) combined with the mean predictions to construct prediction boundaries. This will ensure appropriately calibrated uncertainty estimates of forecast values for operational decisions, allowing the forecasters/stakeholders to select an appropriate threshold for the specific operational application.

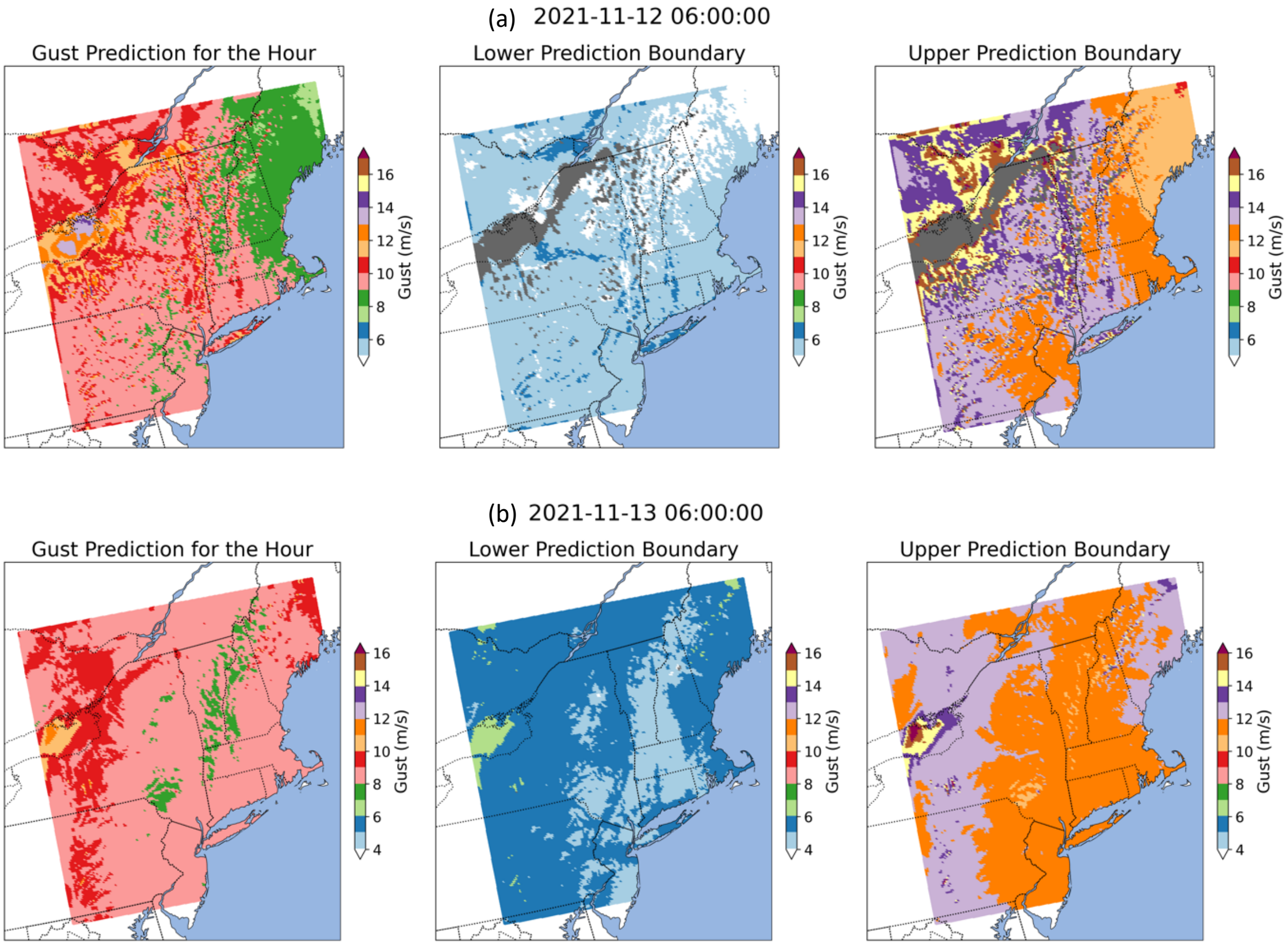

**Fig. 8.** Examples of gust predictions and their respective lower and upper prediction boundaries at (a) 2021-11-12 06 UTC (peak storm intensity) and (b) 2021-11-13 06 UTC (post-peak storm intensity) for the test storm in November of 2021.

We present an example of how gust predictions and their associated uncertainties from ENN can be visually represented for operational forecast applications (Fig. 8). The lower and upper prediction boundaries were computed at 95% confidence level using Eq. 2 and Eq. 3, respectively. For this storm, the 95$^{th}$ percentile of total uncertainty (2.89 m/s) was used as a threshold, with uncertainty in predictions exceeding this threshold labeled as highly uncertain and unreliable. These areas are shown in gray in the middle and rightmost subplots. In Fig. 8a, more gray regions indicate high uncertainty during the storm's peak to construct reasonable prediction boundaries at those regions, whereas Fig. 8b shows the gray region is limited to a smaller part of the White Mountains in New Hampshire only. If we look at the state of Connecticut, the gust prediction for the peak of the storm was around 9-11 m/s (~20-25 mph) (Fig. 8a left), and the minimum and maximum expected gusts were in the range of ~5-7 m/s (Fig. 8a middle) and ~12-15 m/s (Fig. 8a right) (higher along the Connecticut river

valley).These visualizations can offer valuable insights for meteorologists and stakeholders, allowing them to assess forecast ranges and identify situations when forecast boundaries may be less reliable.

### *4.2. Spatial and temporal analysis of estimated uncertainties*

We performed spatial and temporal analysis of estimated uncertainties to investigate how uncertainty evolved with storm intensity. Storm intensity is measured by the maximum surface wind speed or the minimum sea level pressure. Since minimum sea level pressure was not used as a feature in our modeling approach, we used maximum surface wind speed as an indicator for storm intensity. For each hour, we identified the location of the maximum surface wind speed in a test storm across the WRF domain. Similarly, we recorded the location of the spatial maximum total uncertainty predicted by ENN for each hour. Fig. 9 demonstrates that the locations of the spatial maximum total uncertainty aligned with those of the spatial maximum surface wind speed for most hours during the storm. The White Mountain National Forest region in New Hampshire and the Catskills Mountain area in New York consistently showed high total uncertainty in all test storms, except for the 2020-12-05 00 UTC event. These mountain regions have been highlighted in the top leftmost subplot of Fig. 9.

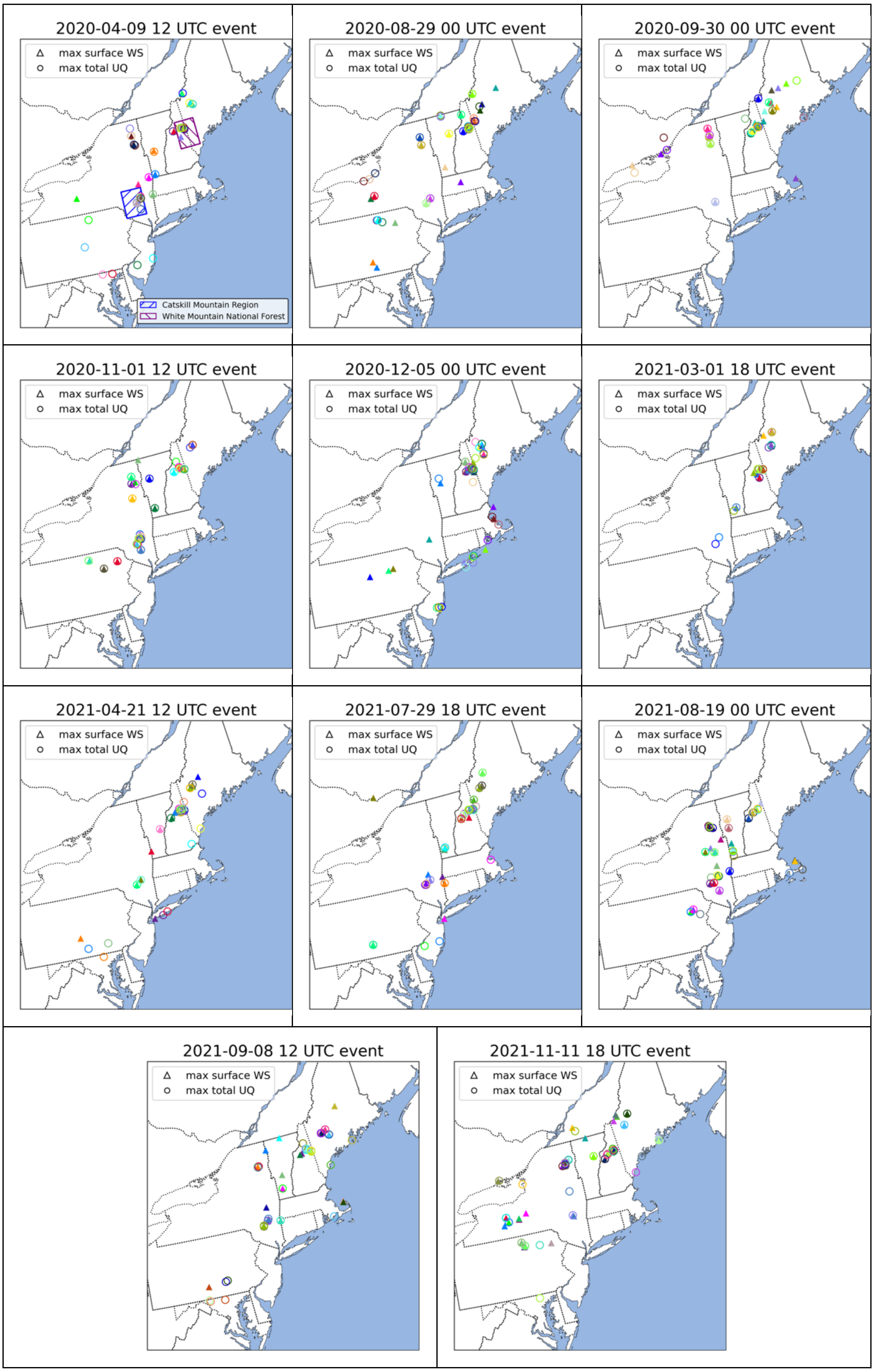

**Fig. 9.** Maps of spatial maximum surface wind speed (triangles) and maximum total uncertainty (total UQ, circles) for each test storm. Each map contains 49 pairs of wind speed and uncertainty values, representing the spatial maxima at each hour over the storm duration (including both the start time and end time resulting in 49 hourly records over a 48-hour period). Each subplot title shows the starting hour of the storm. A circle and a triangle with the same color correspond to the same hour. The White

Mountain National Forest region and the Catskill Mountain have been hatched in the top leftmost subplot as these regions showed maximum total uncertainty consistently over the test storms.

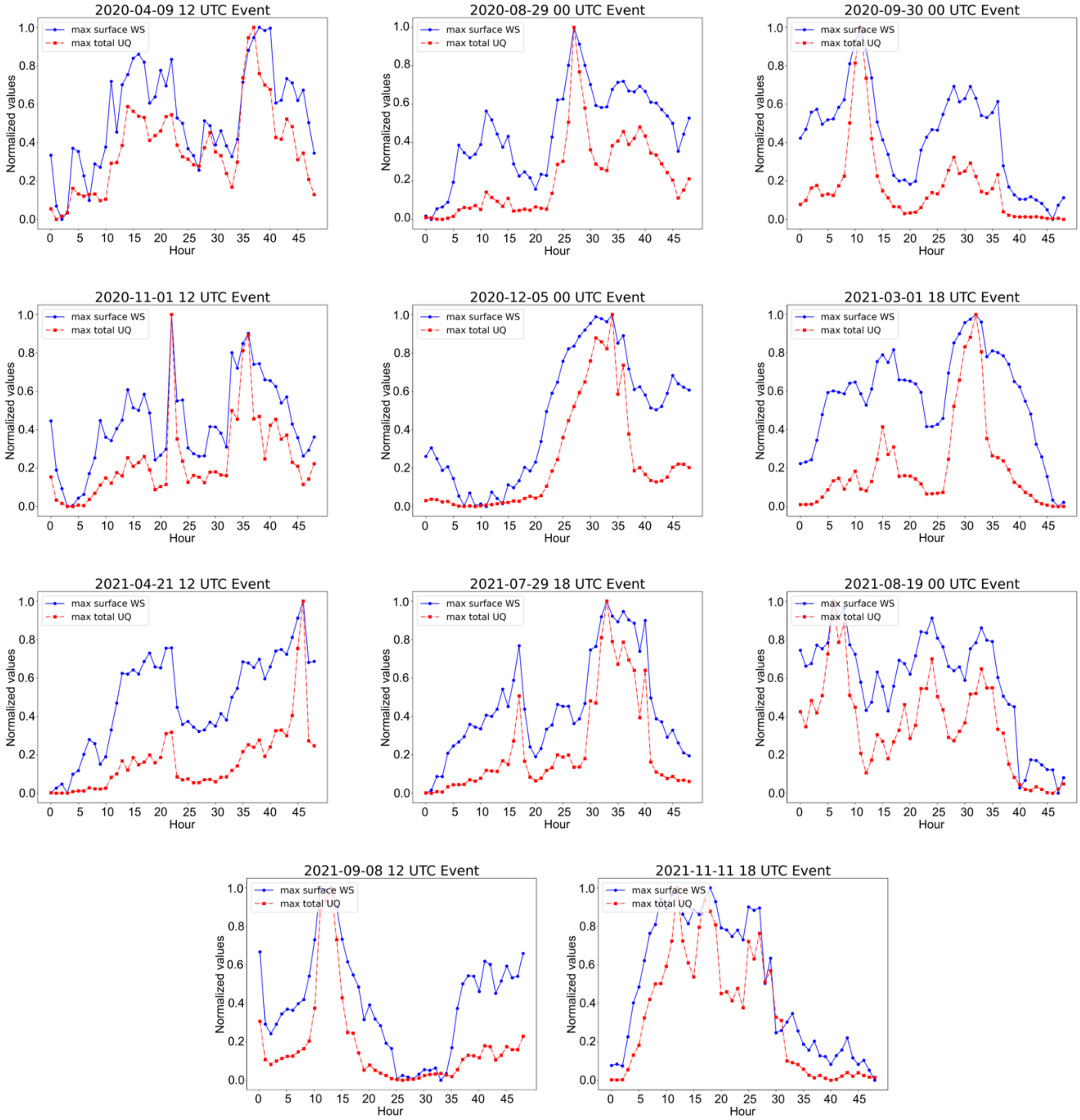

**Fig. 10.** Time Series of spatial maximum surface wind speed and maximum total uncertainty (total UQ) spanning the full duration of each storm. Each variable was normalized as follows: $\frac{V-V_{min}}{V_{max}-V_{min}}$ where $V$= values of the variable, $V_{min}$= minimum value of the variable, $V_{max}$= maximum value of the variable.

The temporal patterns of spatial maximum total uncertainty and spatial maximum surface wind speed also aligned very well, reaching their peaks at the same time (Fig. 10). These results suggest that the majority of total estimated uncertainty and storm intensity were spatially and temporally synchronized. Therefore, the times and regions of peak storm intensity can be expected to experience the highest deviations from mean predictions. By analyzing the uncertainty values and their maxima, forecasters can identify regions of concern and prepare for a wider range of possible outcomes in high-intensity zones.

To further understand how uncertainty varied with predicted gusts, we analyzed the spatial distribution of gust predictions and the total uncertainty in predictions by taking the average of both variables over the storm duration at each grid cell. The analysis revealed that uncertainty in predictions was higher at locations with high gusts. This was expected, since the model becomes less confident when predicting extreme gust values due to their sparsity in the training dataset resulting in high epistemic uncertainty in predictions. Interestingly, we also observed high uncertainty in low gust predictions in some cases which led us to map the spatial gradients of predicted gusts. Fig. 11 shows wind gust predictions (left panels), total uncertainty in predictions (middle panels) and spatial gradients of predicted gusts (right panels) for a subset of the test storms. The spatial gradient of predicted gusts at each grid point was calculated using the formula: $gradient = \frac{|G_{neighbor} - G_{grid}|}{\|S_{neighbor} - S_{grid}\|_2}$ where $G_{neighbor}$ is the gust value at each of the four nearest grid cells, $G_{grid}$ is the gust value at the current grid point, and $\|S_{neighbor} - S_{grid}\|_2$ represents the Euclidean distance between the spatial coordinates of the neighboring grid cells $S_{neighbor}$ and the current grid point $S_{grid}$ defined as: $\|s_{neighbor} - s_{grid}\|_2 = \sqrt{(lon_{neighbor} - lon_{grid})^2 + (lat_{neighbor} - lat_{grid})^2}$. The absolute values of the four gradients were averaged to obtain the final spatial gradient at each grid point.

The storm on 2020-09-30 00 UTC showed high gust predictions along the coastlines (marked as 1 in Fig. 11a left) and low gust predictions near the White Mountain National Forest area (marked as 2 in Fig. 11a left). We observed lower uncertainty in area 1 with uniform spatial gradients of predicted gusts and higher uncertainty in area 2 with relatively higher spatial gradients of predicted gusts (Fig. 11a middle and right). Similar pattern was observed in other storms, for example, the storm on 2020-12-05 00 UTC showed comparatively lower uncertainty along the coast bounded by the dotted line (Fig. 11b middle) where predictions were relatively higher (Fig. 11b left) but the spatial gradients of the predictions were mostly uniform (Fig.11b right). The storms on 2021-08-19 00 UTC and 2021-09-08 12 UTC showed higher uncertainty in areas 1, 2 and 3 (middle panels in Fig. 11c and 11d) despite low gusts (left panels in Fig. 11c and 11d) that can be attributed to the higher spatial gust gradients in those areas (right panels in Fig. 11c and 11d). These results suggest that high uncertainty in predictions was related not only to high gust values but also to high spatial gust gradients.

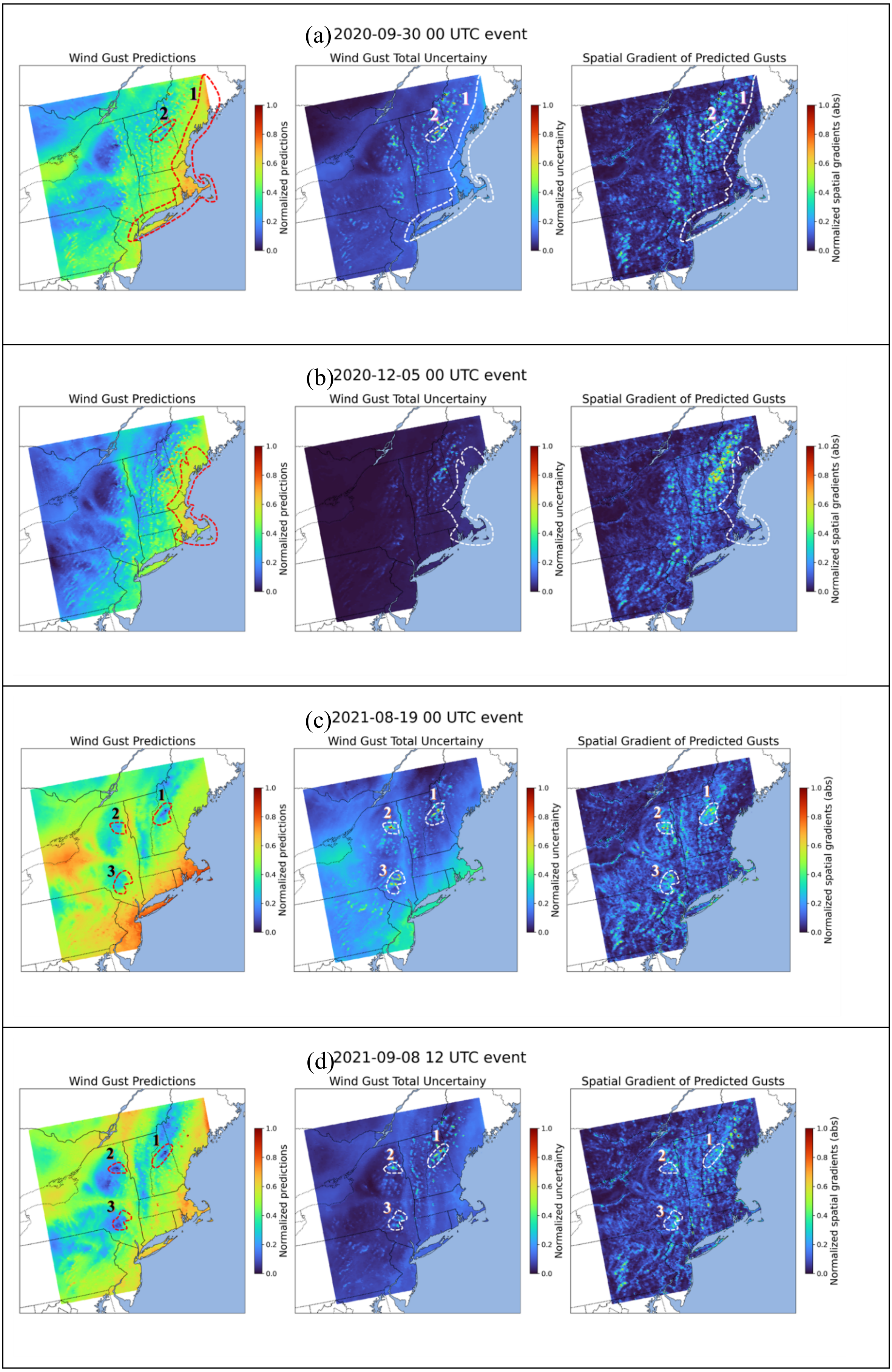

(a) 2020-09-30 00 UTC event
Wind Gust Predictions
Wind Gust Total Uncertainty
Spatial Gradient of Predicted Gusts
Normalized predictions
Normalized uncertainty
Normalized spatial gradients (abs)
(b) 2020-12-05 00 UTC event
Wind Gust Predictions
Wind Gust Total Uncertainty
Spatial Gradient of Predicted Gusts
Normalized predictions
Normalized uncertainty
Normalized spatial gradients (abs)
(c) 2021-08-19 00 UTC event
Wind Gust Predictions
Wind Gust Total Uncertainty
Spatial Gradient of Predicted Gusts
Normalized predictions
Normalized uncertainty
Normalized spatial gradients (abs)
(d) 2021-09-08 12 UTC event
Wind Gust Predictions
Wind Gust Total Uncertainty
Spatial Gradient of Predicted Gusts
Normalized predictions
Normalized uncertainty
Normalized spatial gradients (abs)

**Fig. 11.** Normalized wind gust predictions (left), total uncertainty (middle) and spatial gradients of predicted gusts (right) for a subset of test storms. Predicted gusts and total uncertainty were averaged over the duration of each storm at each grid cell. Each variable was normalized as follows: $\frac{V-V_{min}}{V_{max}-V_{min}}$ where $V$= values of the variable, $V_{min}$= minimum value of the variable, $V_{max}$= maximum value of the variable.

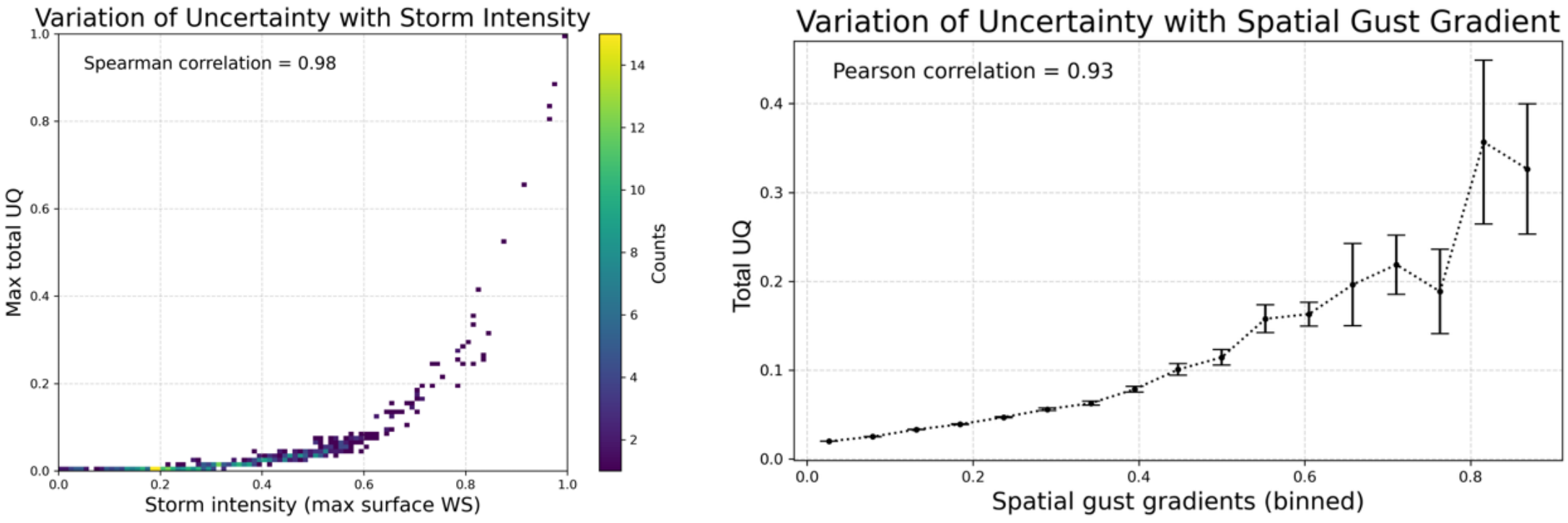

**Fig. 12.** (left) Two-dimensional distribution of spatial maximum total uncertainty (total UQ) and storm intensity across all test storms. (right) Total uncertainty as a function of binned spatial gust gradients for the storms shown in Fig. 11. Gradient values were divided into equal-width bins, and the mean total uncertainty with standard error is shown for each bin. All variables were normalized following the same approach as Fig. 10 and Fig. 11.

Fig. 12 shows the relationships between storm intensity and total uncertainty, as well as between spatial gust gradients and total uncertainty, using correlation-based analyses. The left sublot indicates a strong positive monotonic relationship between storm intensity and spatial maximum uncertainty with a Spearman rank correlation coefficient of 0.98. This suggests that higher surface wind speeds (or storm intensity) were associated with increased predictive uncertainty across storms. The right subplot shows that uncertainty systematically increased with spatial gust gradients with a Pearson correlation coefficient of 0.93, indicating that regions of stronger spatial gust gradients tended to have greater model uncertainty. These findings further support our hypotheses that both higher storm intensity and larger spatial gust gradients likely contribute to increased model uncertainty in gust predictions, however, observational correlations alone do not confirm causality.

### *4.3. Explainability (XAI) of evidential model predictions and uncertainties*

To assess the influence of features on predicted wind gusts and the uncertainty quantities, we used the permutation feature importance (PFI). Features were randomly shuffled 10 times and the error metrics were computed to assess the impact on model performance. We used two metrics in our analysis of PFI: RMSE and the coefficient of determination between the computed RMSE and total uncertainty (R2_RMSE_$\sigma_{total}$) in the spread-skill diagram. An increase in RMSE indicated that the feature negatively impacted the model's prediction of the target variable (wind gust), while a decrease in R2_RMSE_$\sigma_{total}$ signaled a deterioration in model calibration with respect to total uncertainty when the feature was shuffled. The higher the change in the metrics due to shuffling a feature, the more significant that feature's contribution is to the model.

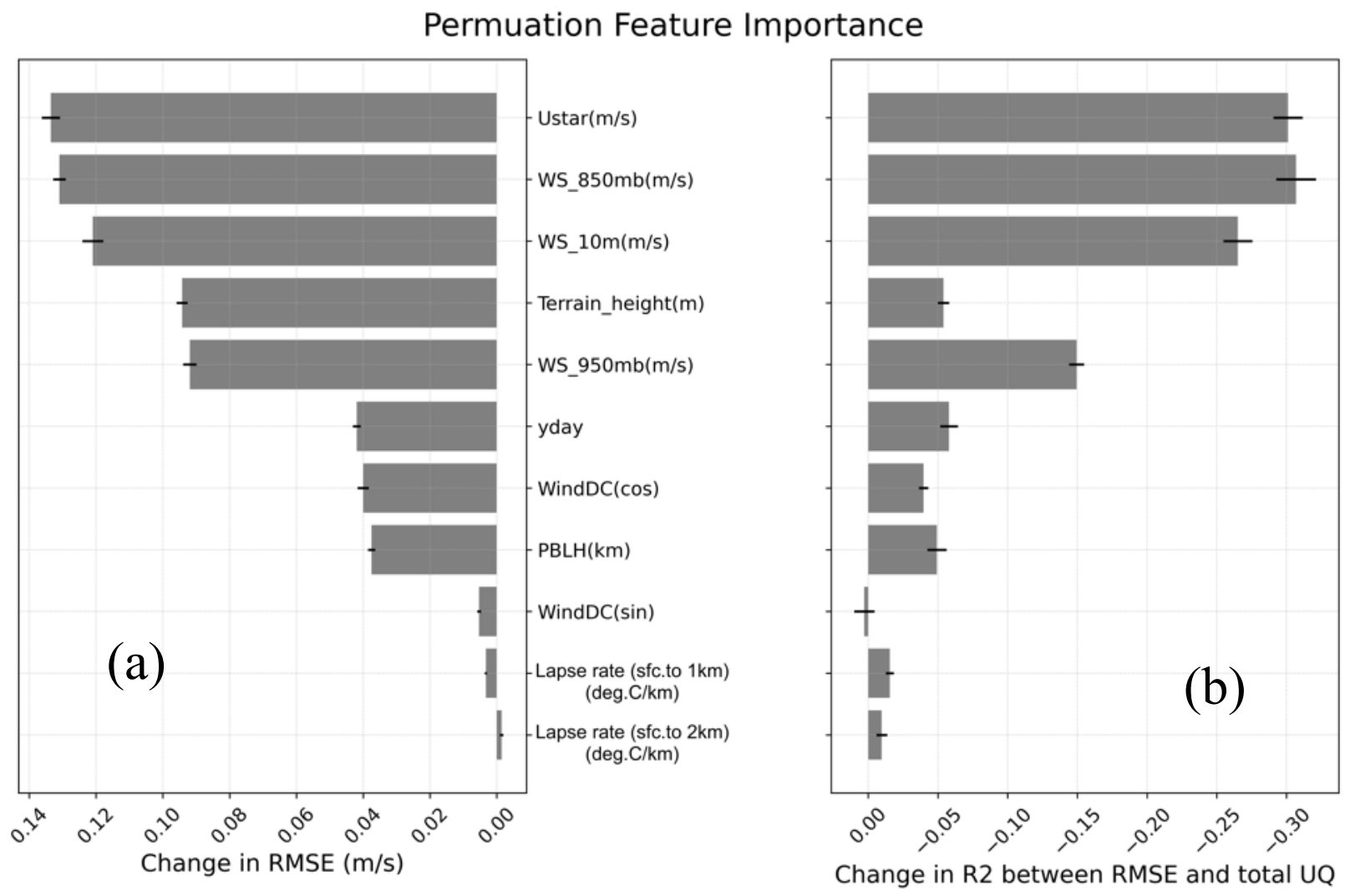

**Fig. 13**. PFI for (a) the model's prediction of wind gusts and (b) the model calibration with respect to total uncertainty (total UQ). The gray bars indicate the average change in the metrics after shuffling a feature 10 times, while the black lines represent the standard deviation of the change in the metrics. PFI was done on the test storms.

Fig. 13 (a) shows that friction velocity, wind speed at different levels and terrain height were the most important predictors for wind gusts. In contrast, cosine-transformed day of the year, cosine of wind direction and planetary boundary layer height were of relatively lower importance, with lapse rates and sine of wind direction having the least impact. A similar ranking is observed in Fig. 13 (b), suggesting that the same top predictors for wind gust predictions also significantly influenced the model calibration with respect to total uncertainty.

Next, we used partial dependence plots (PDPs) to examine the sensitivity of the evidential model to the predictors, i.e., how the model output changed with each feature over its full range. To make a PDP, each feature was divided into 100 equally spaced values across its range. The original feature (e.g., WS_10m) values were systematically replaced with each of these 100 equally spaced values one at a time while other features were kept unchanged. Model predictions were made on this modified dataset and then averaged to obtain the mean prediction and standard deviation at each of these values. It is important to note that PDP reflects the isolated impact of each feature on the model, independent of the distribution of the feature in the original test data (gray histograms on Fig. 14). This is because PDP construction relies on uniform sampling over the range of each feature of interest to provide insight on its effect on model output regardless of data density. Fig. 14 shows two types of PDPs, one for gust prediction (PDP-gust) and the other for uncertainty estimation (PDP-UQ), for wind speed at 10 m, terrain height and lapse rate (surface to 1 km height). PDPs for the remaining features can be found in the Supplement (Fig. S6, S7).

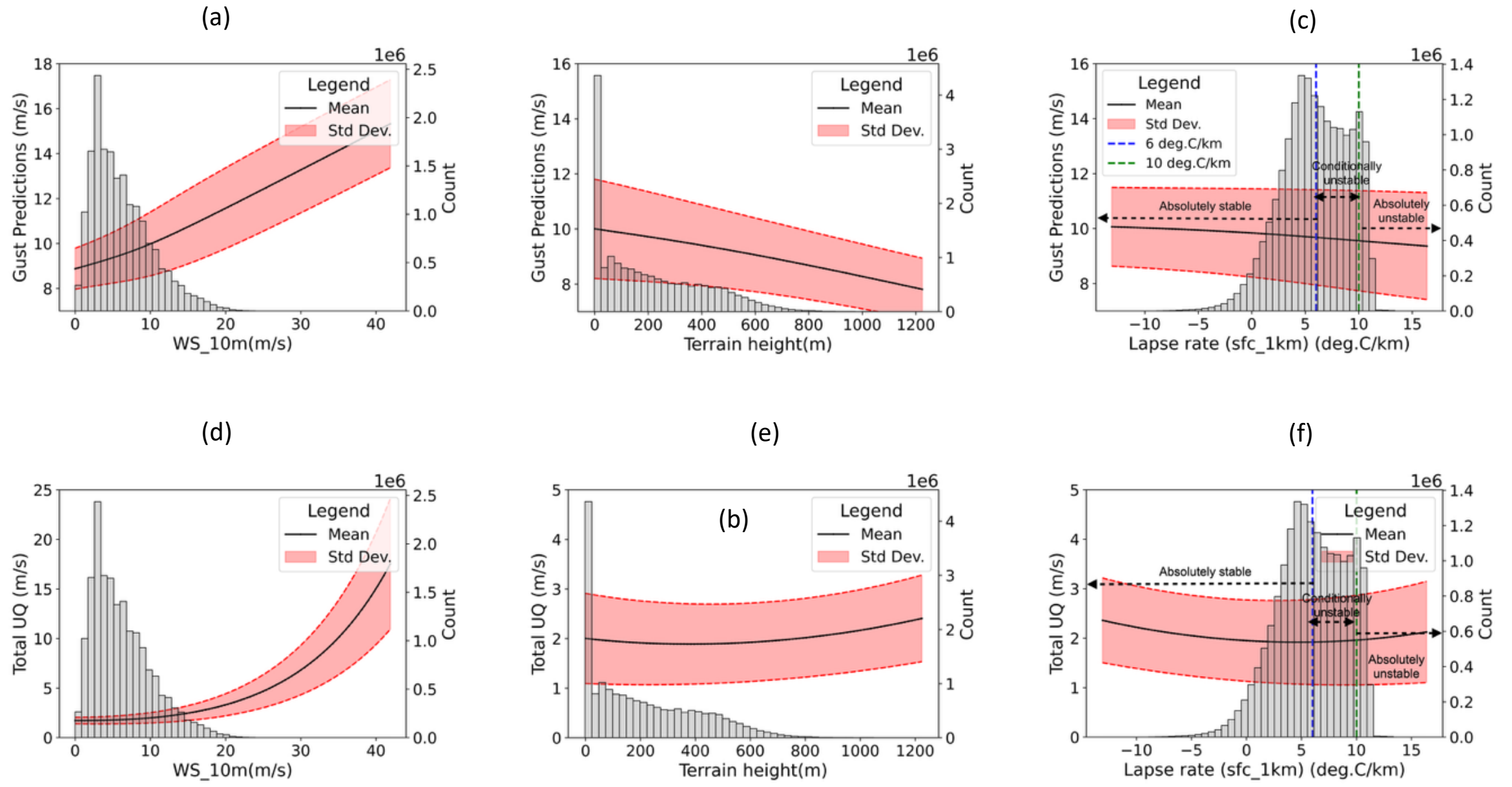

**Fig. 14.** (top row) PDPs for the predicted wind gusts over the test storms. The solid black line represents the mean gust predictions over each feature's full range in the test dataset and the shaded red area was constructed by taking the standard deviation of the predictions. Grey histograms show the distribution of the feature in the test data. (bottom row) PDPs for total uncertainty over the test storms for the same features as in the top row.

With higher surface wind speeds, gust predictions increased (Fig. 14a) as expected but decreased with higher terrain height (Fig. 14b) contrary to the expectation of stronger gusts at higher altitudes. This likely resulted from an imbalance in the training data, with fewer high-altitude stations reporting strong gusts compared to lower-altitude stations with high gusts (Fig. S9, Supplement). PDP-gust for the lapse rate (surface to 1 km height) showed a gradual decline of predicted gusts as the atmosphere shifted from stable to unstable (Fig. 14c). A stable atmosphere typically results in limited vertical mixing and thus lower winds, while an unstable atmosphere promotes vertical mixing where high winds and gusts are expected. Thus, the expected relationship between wind gusts and atmospheric stability was not entirely upheld.

PDP-UQ for surface wind speed (Fig. 14d) showed an initial flattening in total uncertainty, followed by an exponential increase to unrealistic values. Such inflated values should only be interpreted as highly uncertain predictions, rather than assigning them any physical significance. Terrain height showed a subtle decline in total uncertainty before trending upward (Fig. 14e). For lapse rate (surface to 1 km height) (Fig. 14f), uncertainty decreased as the atmosphere transitioned from negative (inversion) to positive, suggesting higher model confidence in stable conditions. However, uncertainty increased gradually as the atmosphere shifted to absolutely unstable condition from conditional instability, reflecting reduced model confidence in turbulent conditions. This complex interplay between atmospheric stability and uncertainty aligns with the inherent chaotic nature during transitions between stability states, where the atmospheric dynamics can be highly variable.

The large spread of the standard deviation (Std Dev.) swaths in Fig. 14 suggests the presence of higher-order interactions among the features. These interactions may not be fully captured in the PDPs due to averaging and are being partially obscured in the visualizations. It is also important to recognize that assessing PDPs for uncertainty remains challenging without knowledge of a ground truth, making it difficult to directly analyze whether the findings align with atmospheric dynamics. Our initial hypothesis was that uncertainty in predictions would be higher for features that were less informative for the model. However, the XAI techniques used in this study did not corroborate the hypotheses. In fact, features ranked highly for gust predictions and model calibration in PFI analysis exhibited higher uncertainty across PDPs, while features with marginal uncertainty increase (Fig. 14d-f and Fig. S7 in the Supplement) had less impact on both the gust prediction and model calibration.

### *4.4. Challenges and limitations to the evidential and XAI approach*

Training the evidential model posed several challenges. For example, despite extensive hyperparameter tuning through ECHO, the evidential coefficient still required some adjustment through trial and error. The process can be time-consuming, especially with the use of multi-objective optimization. Even after the exhaustive hyperparameter search, occasional unrealistic uncertainty values appeared. However, the distribution of the total uncertainty (Fig. 5) showed that the uncertainty estimates of the model were reasonable at the 95$^{th}$ percentile, which is an acceptable threshold in many applications. Another limitation is the evidential coefficient's sensitivity to data. A shift in the underlying data might alter the model calibration with respect to uncertainty, though the predictions of the target variable remain relatively stable. Such sensitivity of the evidential coefficient to changes in underlying data may necessitate additional tuning for proper calibration when applied to different regions or storm types, thus potentially limiting the model's generalizability. In a future study, we will focus on developing techniques to constrain uncertainty values within reasonable thresholds, ensuring they remain consistent with physical boundaries of the target variable.

The XAI analyses were performed on aggregated data from all test storms. Therefore, the results reflect the overall feature importance and partial dependence of predicted gusts and uncertainty estimation when a group of similar storms are combined. We plan to include storm-specific and region-specific XAI analyses in a future study. It is important to note that these XAI techniques, e.g., PDPs though easy to implement and intuitive to comprehend, come with their own challenges and must be interpreted with caution. PDPs assume feature independence, which is not true for all features used in this study. In addition, if a feature has a heterogeneous relationship with the target (e.g., some data points may have a positive and some may have a negative association), the PDPs will mask those out since they provide an average effect (Greenwell et al. 2018; Molnar, 2022; Zhao and Hastie, 2019).

## 5. Concluding Remarks

Evidential deep learning offered an efficient approach to uncertainty estimation for predicted wind gusts without the need to run an ensemble. While it reduced the overprediction of gusts from the physics-based WRF model, it struggled to address the underprediction of high gusts - a limitation also observed in the XGBoost model, as noted in J24. The difficulty of both NN and XGBoost models in capturing extreme values in right-skewed distributions, such as wind gusts, underscores the importance of uncertainty quantification.

Epistemic uncertainty was approximately three times greater than aleatoric uncertainty, making it the primary contributor to total uncertainty in gust predictions for the extratropical storms used in this study. We evaluated the total uncertainty estimates of the test storms by constructing prediction intervals at various confidence levels, excluding inflated and unrealistic uncertainty values (> $95^{th}$ percentile) to prevent biased optimistic outcomes in the PICP analysis. At 95% confidence level, 179 stations out of 266 stations captured 95% or more of the observed gusts within the prediction intervals. Comparison of PICP values at different confidence thresholds demonstrated that defining prediction boundaries using factored total uncertainty, rather than simply adding/subtracting it from the mean prediction, helped prevent overconfidence in predictions. From an operational perspective, providing gust forecasts by defining uncertainty in the predictions is a crucial step in building stakeholders' confidence for assessing risk, planning and responding to extreme gust events.

Spatial and temporal analysis of maximum hourly surface wind speed, used as a proxy for storm intensity, showed that uncertainty estimates closely aligned with storm intensity for all test storms. Uncertainty was consistently high in regions like White Mountain National Forest and the Catskill Mountains, suggesting caution in interpreting uncertainty in mountainous areas. Spatial analysis also revealed that in addition to high gust predictions, areas with higher spatial gust gradients exhibited higher uncertainty.

To understand the sensitivity of model output across features, we used PFI and PDPs. PFI demonstrated a similar feature ranking for both gust prediction and model calibration in terms of total uncertainty. Contrary to our initial hypothesis that less informative predictors would yield higher uncertainty, PDPs revealed that the most important predictors resulted in greater variation in uncertainty values, whereas less important predictors exhibited minimal variation. This suggests that while the evidential approach effectively differentiates between more confident and less confident predictions, it is not recommended to reconfigure the model by eliminating predictors that show greater variation in uncertainty, as doing so may degrade the model's performance.

Lastly, while ENN has the potential to improve gust predictions and UQ as a post-processing tool, some limitations remain. Same as previous studies related to ENN, we observed occasional unrealistic uncertainty values beyond the physical bounds of the target variable, warranting careful interpretation of these values. In our analysis, we used the 95$^{th}$ percentile of total uncertainty as a threshold to construct prediction boundaries, however, this choice is subjective, and stakeholders may empirically establish their own thresholds depending on the specific needs and objectives of their work. Another challenge is the need for extensive hyperparameter tuning, especially for the evidential coefficient, which primarily affects uncertainty calibration rather than prediction accuracy of the target variable. Further research is needed to constrain uncertainty values within reasonable limits to prevent them from exceeding the computed error by a significant margin.

**CRediT authorship contribution statement**

**I. Jahan:** Writing – original draft, Writing – review & editing, Conceptualization, Data curation, Formal analysis, Methodology, Software, Validation, Visualization. **J. S. Schreck:** Writing – review & editing, Conceptualization, Methodology, Software, Supervision, Resources, Funding acquisition. **D. J. Gagne:** Conceptualization, Methodology, Software, Resources, Funding acquisition. **C. Becker:** Writing – review & editing, Conceptualization, Methodology, Software, Resources. **M. Astitha:** Writing – review & editing, Conceptualization, Supervision, Resources, Funding acquisition, Project Administration.

**Software and Data Availability**

All datasets and codes used in this study are available in the Open Science Framework (OSF): https://osf.io/ryd6x/?view_only=74af2189d1ed49c290eccbff5e9987a5. The repository also includes documentation for the codes and datasets. It was created by Marina Astitha (E-mail: marina.astitha@ uconn.edu) in 2025. Training and validation of ENN were performed using the MILES-GUESS python package available in the GitHub repository: https://github.com/ai2es/miles-guess/tree/main/mlguess. All work related to this study was conducted on a high-performance computer with Linux operating system.

**Declaration of interests**

The authors declare no competing financial interests or personal relationships that could have influenced the work reported in this paper.

**Declaration of generative AI and AI-assisted technologies in the writing process**

During the preparation of this work the author(s) used ChatGPT-4o to improve readability of the manuscript. After using this tool, the author(s) reviewed and edited the content as needed and take(s) full responsibility for the content of the publication.

**Acknowledgments**

This work was partially supported by the GE Fellowship for Excellence awarded to author Israt Jahan. This work was also partially supported by the Eversource Energy Center at the University of Connecticut through the research grant "Improving Extreme Weather Forecasting Capabilities in support of Power Outage Prediction Activities: Phase II – wind gust and winter weather" awarded to Prof. Astitha. This research is made possible by the New York State (NYS) Mesonet. Original funding for the NYS Mesonet (NYSM) buildup was provided by Federal Emergency Management Agency grant FEMA-4085-DR-NY. The continued operation and maintenance of the NYSM is supported by the National Mesonet Program, University at Albany, Federal and private grants, and others. We acknowledge high-performance computing support from the Derecho (doi:10.5065/qx9a-pg09) and the Casper system (**https://ncar.pub/casper**) provided by the NSF National Center for Atmospheric Research (NCAR), sponsored by the National Science Foundation (NSF). We extend our gratitude to the NSF NCAR CISL for hosting author Israt Jahan through the CISL Visitor Program, which contributed to the development of this manuscript.

## APPENDIX A

### List of storms

**Table A1.** Dates of the 61 storms used in this study. Storms 1-42 = training (T); Storms 43-50 = validation (V); Storms 51-61 = testing(S). Each storm had a 48-hour duration. The WRF simulation start date was 12 hours before the storm start date.

| | **Storm Start Date** | **Time (UTC)** | | **Storm Start Date** | **Time (UTC)** | | **Storm Start Date** | **Time (UTC)** |
|---|---|---|---|---|---|---|---|---|
| **1-T** | 01/01/2017 | 0 | **21-T** | 07/29/2017 | 0 | **41-T** | 01/09/2019 | 0 |
| **2-T** | 01/04/2017 | 18 | **22-T** | 08/18/2017 | 6 | **42-T** | 02/08/2019 | 0 |
| **3-T** | 01/10/2017 | 6 | **23-T** | 08/29/2017 | 12 | **43V** | 02/12/2019 | 12 |
| **4-T** | 01/17/2017 | 6 | **24-T** | 09/03/2017 | 0 | **44-V** | 02/20/2019 | 12 |
| **5-T** | 01/23/2017 | 6 | **25-T** | 09/05/2017 | 12 | **45-V** | 02/25/2019 | 0 |
| **6-T** | 02/01/2017 | 12 | **26-T** | 10/09/2017 | 0 | **46-V** | 10/16/2019 | 12 |
| **7-T** | 03/10/2017 | 18 | **27-T** | 10/15/2017 | 6 | **47-V** | 10/31/2019 | 12 |
| **8-T** | 03/16/2017 | 6 | **28-T** | 10/24/2017 | 0 | **48-V** | 01/11/2020 | 6 |
| **9-T** | 04/12/2017 | 12 | **29-T** | 10/29/2017 | 12 | **49-V** | 02/07/2020 | 0 |
| **10-T** | 04/16/2017 | 12 | **30-T** | 11/06/2017 | 6 | **50-V** | 03/03/2020 | 18 |
| **11-T** | 04/25/2017 | 6 | **31-T** | 11/10/2017 | 0 | **51-S** | 04/09/2020 | 12 |
| **12T** | 04/29/2017 | 0 | **32-T** | 12/22/2017 | 18 | **52-S** | 08/29/2020 | 0 |
| **13-T** | 05/02/2017 | 0 | **33-T** | 01/04/2018 | 6 | **53-S** | 09/30/2020 | 0 |
| **14-T** | 05/05/2017 | 6 | **34-T** | 02/07/2018 | 6 | **54-S** | 11/01/2020 | 12 |
| **15-T** | 05/13/2017 | 0 | **35-T** | 02/17/2018 | 18 | **55-S** | 12/05/2020 | 0 |
| **16-T** | 05/18/2017 | 6 | **36-T** | 10/27/2018 | 6 | **56-S** | 03/01/2021 | 18 |
| **17-T** | 05/25/2017 | 6 | **37-T** | 11/03/2018 | 0 | **57-S** | 04/21/2021 | 12 |
| **18-T** | 06/05/2017 | 6 | **38-T** | 11/06/2018 | 12 | **58-S** | 07/29/2021 | 18 |
| **19-T** | 06/16/2017 | 0 | **39-T** | 11/15/2018 | 18 | **59-S** | 08/19/2021 | 0 |
| **20-T** | 07/24/2017 | 0 | **40-T** | 01/01/2019 | 0 | **60-S** | 09/08/2021 | 12 |
| | | | | | | **61-S** | 11/11/2021 | 18 |

# Supplementary Document
## for the paper
# Uncertainty Quantification of Wind Gust Predictions in the Northeast US: An Evidential Neural Network and Explainable Artificial Intelligence Approach

Israt Jahan[1], John S. Schreck[4], David John Gagne[3,4], Charlie Becker[4], Marina Astitha[1,2,3]

[1]*School of Civil and Environmental Engineering, University of Connecticut, Storrs, CT, USA*

[2]*Eversource Energy Center, University of Connecticut, Storrs, CT, USA*

[3]*NSF National Center for Atmospheric Research (NCAR), Research Applications Laboratory, Boulder, CO, USA*

[4]*NSF National Center for Atmospheric Research (NCAR), Computational & Information Systems Lab, Boulder, CO, USA*

*Corresponding author*: Marina Astitha, marina.astitha@uconn.edu

## 1. Evidential Regression and Law of Total Variance

For a regression dataset D = $\{x_i, y_i\}_{i=1}^{N}$, where the target values are assumed to be drawn independently and identically distributed from a Gaussian distribution with unknown mean and variance $\theta = \{\mu, \sigma^2\}$, evidential regression provides probabilistic estimates of $\theta$ (Amini et al., 2020; Schreck et al., 2024; Soleimany et al., 2021). In this approach, the mean $\mu$ is assumed to be drawn from a Gaussian distribution while and the variance $\sigma^2$ follows an Inverse-Gamma distribution. The joint higher-order distribution is referred to as the evidential distribution and can be expressed as a Normal-Inverse-Gamma distribution. The Normal-Inverse-Gamma distribution p($\theta|m$) is parametrized by m = $\{\gamma, \nu, \alpha, \beta\}$, which represents a distribution over $\theta = \{\mu, \sigma^2\}$. In ENN, the final layer is modified to output these four Normal-Inverse-Gamma parameters per target.

According to the law of total variance (LoTV; Casella and Berger, 2002), for two random variables X and Y on the same probability space, the variance of variable Y can be divided into two parts:

$$Var(Y) = E[Var(Y \mid X)] + Var(E[Y \mid X]) \quad \text{(Eq. S1)}$$

where $Var(Y)$ represents the total variance of $Y$ (interpreted as total uncertainty of prediction), $E[Var(Y \mid X)]$ is the expected conditional variance of Y given X (interpreted as aleatoric uncertainty or inherent data uncertainty), and $Var(E[Y \mid X])$ is the variance of the conditional mean of Y given X (interpreted as epistemic uncertainty or model uncertainty). Applying the LoTV to the Normal-Inverse Gamma distribution yields:

$$E[\mu] = \gamma \ (Mean\ prediction) \quad \text{(Eq. S2)}$$

$$E[\sigma^2] = \frac{\beta}{\alpha - 1} \ (Aleatoric) \quad \text{(Eq. S3)}$$

$$Var(\mu) = \frac{\beta}{\nu(\alpha - 1)} \ (Epistemic) \quad \text{(Eq. S4)}$$

From the Normal-Inverse-Gamma parameters, ENN computes the uncertainty components. To summarize, for each data point (or sample), ENN provides a prediction which we refer to as 'mean prediction' since it represents the central value around which the actual target is expected to lie and the variances denoting the uncertainty or spread around the mean. Taking the square root of the variances, we get uncertainty in the same unit as the target variable (e.g., m/s in this study).

ENNs are trained with a dual-objective loss function $L(x)$ comprising two terms: the negative log-likelihood $L^{NLL}(x)$ to maximize model fit and a regularizer $L^R(x)$ to suppress evidence or raise the uncertainty in support of incorrect predictions:

$$L(x) = L^{NLL}(x) + \lambda L^R(x) \quad \text{(Eq. S5)}$$

The regularization coefficient $\lambda$, also known as the evidential coefficient, controls the calibration of the model by adjusting the uncertainty values. More details on the mathematical formulation of evidential regression can be found in Amini et al. (2020) and Schreck et al. (2024).

## 2. Evaluation Metrics

### *2.1 Gust Predictions*

To assess the error in predicted gusts, we used the metrics outlined in Table S1. $\hat{y}_i$ is the predicted gust for i[th] observation, $y_i$ is the observed gust for i[th] observation, and n is the total number of observations. $\overline{\hat{y}}$ is the mean of predicted gusts and $\bar{y}$ is the mean of observed gusts.

**Table S1.** Evaluation metrics used to assess the error in ENN gust predictions.

| Error metric | Equation |
|---|---|
| Mean bias | $\frac{\sum_{i=1}^{n}(\widehat{y_i} - y_i)}{n}$ |
| Mean absolute error (MAE) | $\frac{\sum_{i=1}^{n}\lvert\widehat{y_i} - y_i\rvert}{n}$ |
| Root mean squared error (RMSE) | $\sqrt{\frac{\sum_{i=1}^{n}(y_i - \hat{y}_i)^2}{n}}$ |
| Centered root mean squared error (CRMSE) | $\sqrt{(RMSE^2 - Bias^2)}$ |
| Pearson correlation coefficient (r) | $r = \frac{\sum_{i=1}^{n}(y_i - \bar{y})(\hat{y}_i - \bar{\hat{y}})}{\sqrt{\sum_{i=1}^{n}(y_i - \bar{y})^2}\sqrt{\sum_{i=1}^{n}(\hat{y}_i - \bar{\hat{y}})^2}}$ |

### *2.2 Uncertainty Calibration*

The discard fraction diagram shows the change in RMSE with a systematic elimination of data points with higher uncertainties, allowing for an evaluation of how model performance progressively improves (Barnes and Barnes, 2021) with more confident predictions. We also computed the dependency of RMSE on the predicted spread (or uncertainty), depicted by the spread-skill diagram (Luca Delle Monache et al., 2013). The principle behind this is that uncertainty and RMSE should be proportional, with more uncertain data points expected to result in higher RMSE. A 1-1 relationship in the spread-skill diagram suggests that the model is calibrated according to its uncertainty estimates (Schreck et al., 2024)

The probability integral transform (PIT) represents the quantile of the predicted distribution where the observation falls, computed by evaluating the cumulative distribution function of the predicted distribution at the observed value. A uniform PIT histogram signals a perfectly calibrated model. During hyperparameter tuning, we used PITD skill score as an optimization metric which measures the deviation from uniformity in the PIT histogram relative to the worst possible PITD score (Schreck et al., 2024).

$$PITD = \sqrt{\frac{1}{M}\sum_{m=1}^{M}\left(\frac{N_m}{N} - \frac{1}{M}\right)^2} \quad \text{(Eq. S6)}$$

where $M$ is the number of bins, $Nm$ is the count of samples in each bin, and $N$ is the total number of samples.

$$PITD\ skill\ score\ = 1 - \frac{PITD}{PITD_{\ worst}} \qquad \text{(Eq. S7)}$$

where $PITD_{worst}$ represents the worst possible PITD score, assuming that all predictions end up in one of the bounding bins of the PIT histogram. The PITD skill score ranges from 0 to 1, with 0 representing no calibration and 1 indicating perfect calibration.

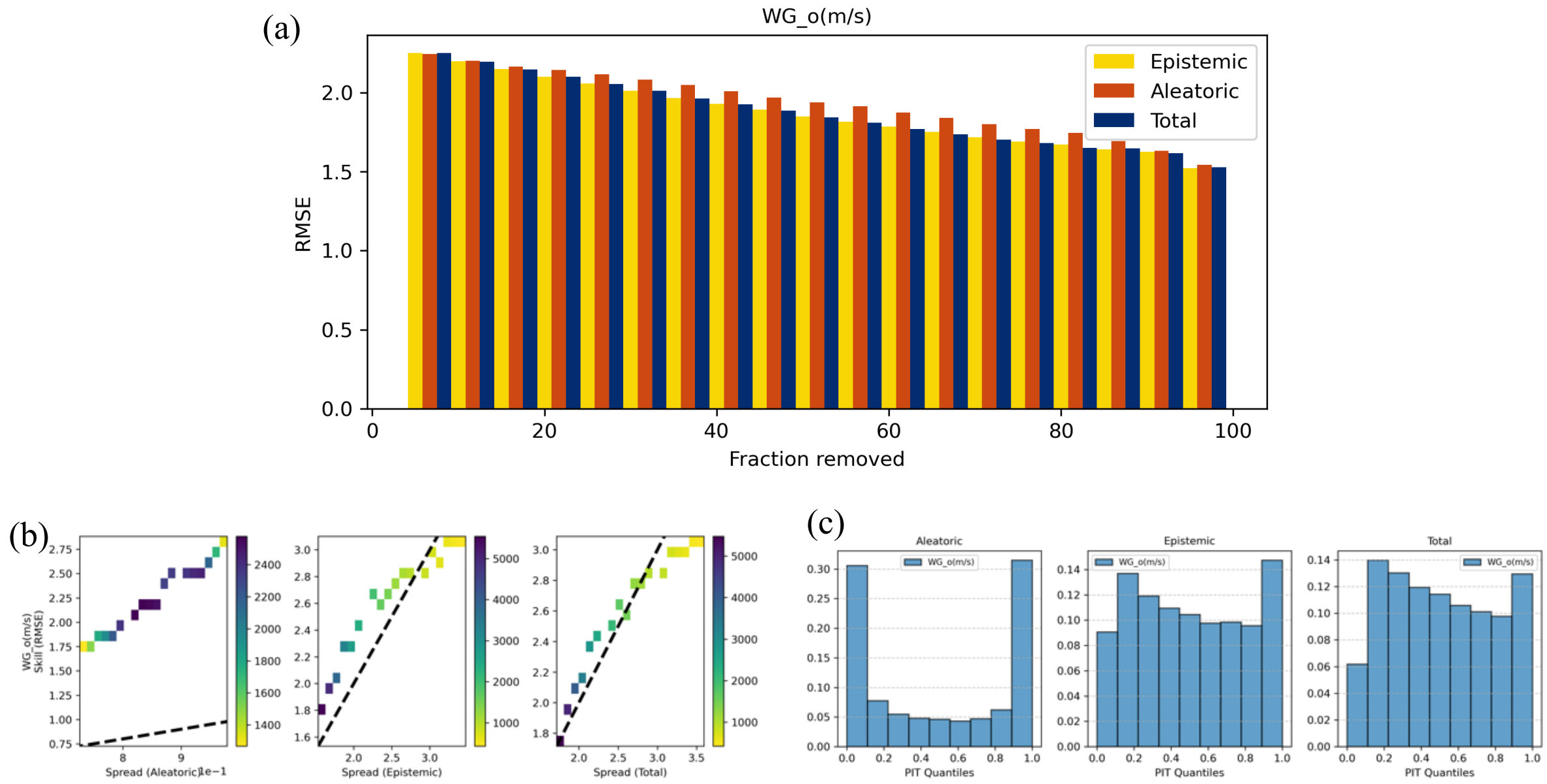

**Figure S1.** (a) The discard fraction diagram illustrates the relationship between the fraction of data points removed and the remaining subset's RMSE. (b) The spread-skill relationship is illustrated using a 2D histogram, illustrating the relationship between the uncertainties [standard deviations ($\sigma$)] and RMSE for the ENN. (c) PIT histograms constructed using the estimated aleatoric, epistemic and total uncertainty. All plots were generated using test data.

## 3. Hyperparameter Tuning

The Earth Computing Hyperparameter Optimization (ECHO) tool (Schreck and Gagne, 2021), built with Optuna, automatically adjusts its hyperparameter search space based on how the trials are performing. Optuna is an automated hyperparameter optimization framework designed to efficiently find optimal hyperparameters through a sequential optimization process, e.g., TPE algorithm. Optuna also supports pruning of unpromising trials, improving both the speed and performance of the optimization process (Akiba et al., 2019). We evaluated model performance using three metrics on the validation data: mean absolute error (val_mae), $R^2$ between the root mean squared error (RMSE) and total uncertainty (val_r2_rmse_sigma total), and the PITD skill score (val_pitd). The objective was to get optimized hyperparameters that would minimize val_mae while maximizing val_r2_rmse_sigma total and val_pitd. The

val_mae metric helps in reducing the prediction error for gusts, while the other two metrics focus on exploring the hyperparameter space to achieve the best model calibration for total uncertainty (Schreck et al., 2024).

**Table S2.** Hyperparameter space and the optimum values identified by ECHO.

| ***Hyperparameter*** | ***Hyperparameter space*** | | ***Optimum value*** |
|---|---|---|---|
| | ***Lowest*** | ***Highest*** | |
| Learning rate | $1 \times 10^{-6}$ | 0.01 | $1.69 \times 10^{-6}$ |
| Dropout alpha | 0 | 0.5 | 0.15 |
| Hidden layers | 1 | 5 | 1 |
| Hidden neurons | 1 | 1000 | 823 |
| Batch size | 10 | 20000 | 2000 |
| Evidential coefficient | $1 \times 10^{-5}$ | 100 | 0.59 |
| l1 weight | $1 \times 10^{-12}$ | 0.01 | $5.14 \times 10^{-11}$ |
| l2 weight | $1 \times 10^{-12}$ | 0.01 | $5.11 \times 10^{-8}$ |

## 4. Data Augmentation

For data augmentation, we used SMOGN (https://github.com/nickkunz/smogn) which is a python implementation of synthetic minority over-sampling technique for regression with gaussian noise (Branco et al., 2017). We attempted SMOGN with three relevance thresholds (RTs) of 0.2, 0.5 and 0.8. We applied the data augmentation only in training data while validation and test data were kept unchanged. Fig. S2 shows the distribution of observed gusts in the original training data and SMOGN augmented training data with three different relevance thresholds.

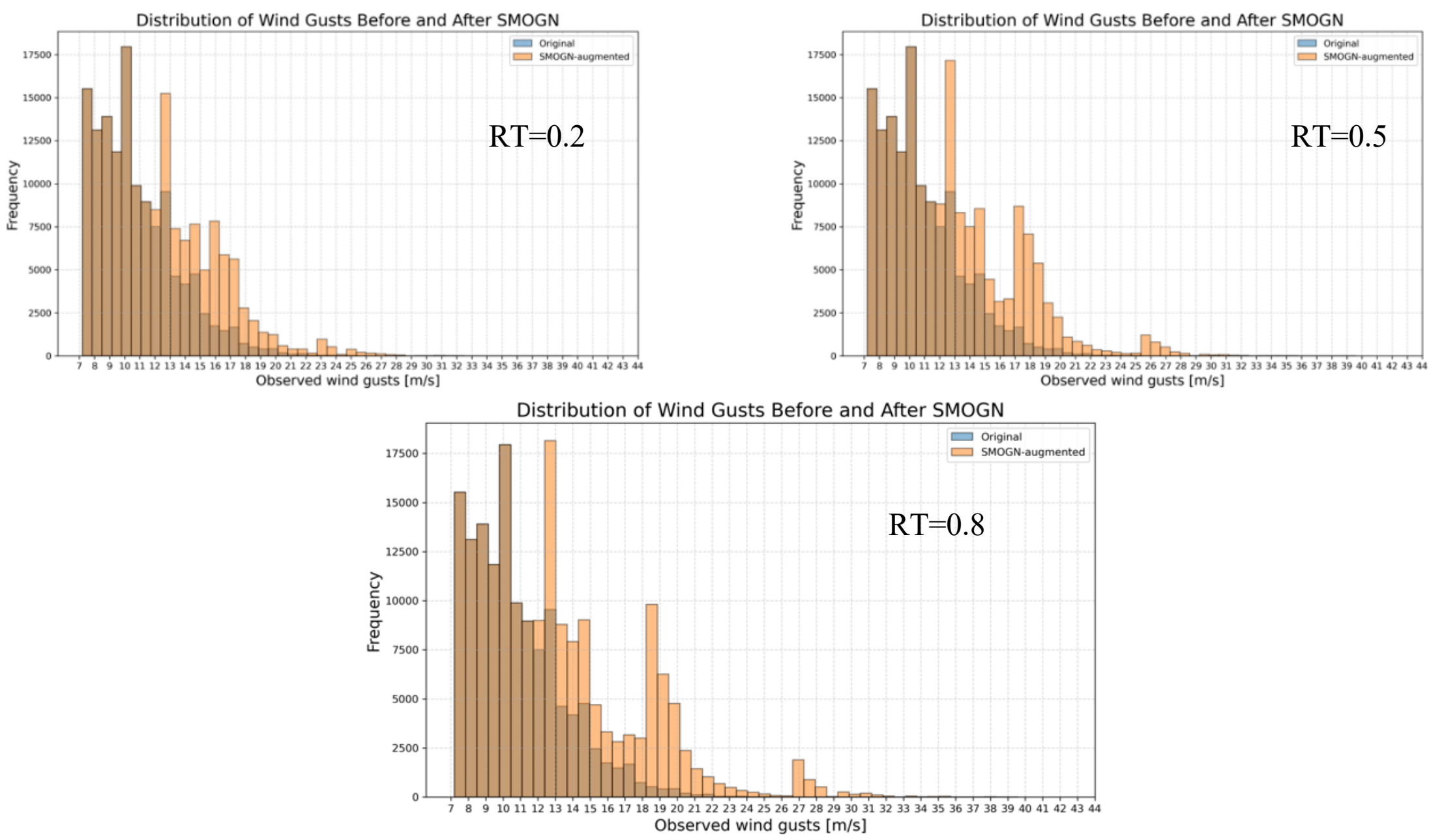

**Figure S2.** Distribution of observed wind gusts in the training data before (original) and after data augmentation (SMOGN-augmented) with relevance thresholds (RTs) of 0.2, 0.5 and 0.8.

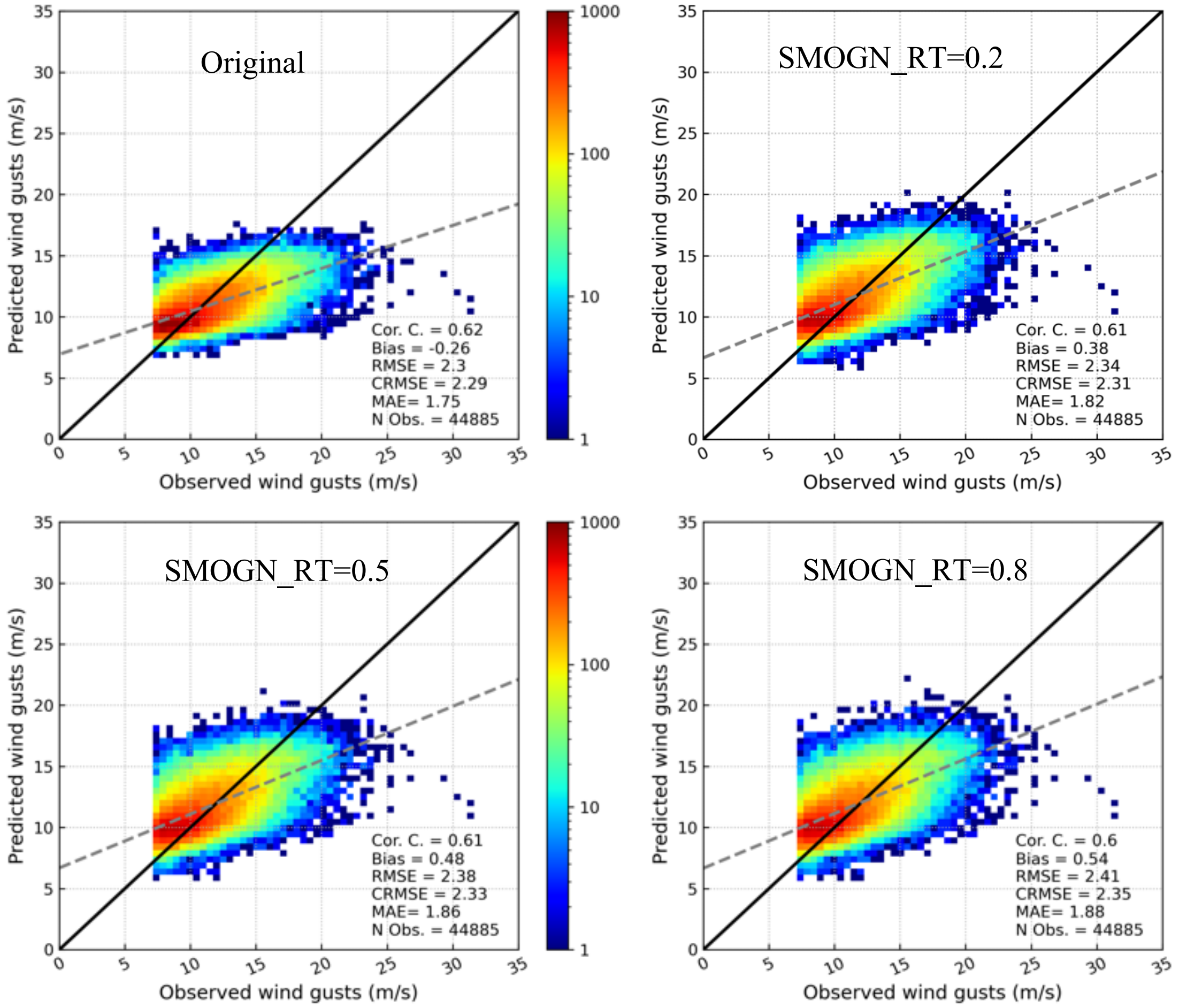

**Figure S3.** (top left) Wind gust predictions from ENN without any data augmentation, (top right) after data augmentation with SMOGN having RT of 0.2, (bottom left) RT of 0.5, and (bottom right) RT of 0.8. Bias, RMSE, CRMSE and MAE have the same units as wind gust, which is m/s. All results are based on test data.

The scatter plots (Fig. S3) for SMOGN show slight changes in the predictions of higher gusts shifting towards the black 1-1 line, however, the overall error statistics was still slightly better in the original ENN. To quantify the change in the model performance, we performed error analysis stratified by observed gust intensity percentiles. We divided the target distribution into 20 percentile bins (0–5%, 5–10%, ..., 95–100%) based on the observed gust values in our data. For each bin, we computed Mean Absolute Error (MAE) and Root Mean Squared Error (RMSE) separately for the ENN model without SMOGN and the three SMOGN-trained ENN models (RT=0.2, RT=0.5 and RT=0.8). The results have been summarized in Table S3.

**Table S3.** Stratified error analysis of wind gust predictions without SMOGN and with SMOGN. The error metrics without SMOGN have been labeled as 'original'. Gust percentile thresholds, MAE and RMSE are in m/s. The error analysis was conducted on test data. The lowest MAE and RMSE values are in bold fonts.

| Bin | Gust percentile threshold | Count | MAE ENN original | MAE SMOGN RT 0.2 | MAE SMOGN RT 0.5 | MAE SMOGN RT 0.8 | RMSE ENN original | RMSE SMOGN RT 0.2 | RMSE SMOGN RT 0.5 | RMSE SMOGN RT 0.8 |
|---|---|---|---|---|---|---|---|---|---|---|
| 0-5 | [7.2, 7.6) | 2072 | **2.17** | 3.32 | 3.40 | 3.44 | **2.38** | 3.75 | 3.84 | 3.89 |
| 5-10 | [7.6, 7.8) | 2185 | **2.05** | 3.01 | 3.11 | 3.15 | **2.29** | 3.50 | 3.61 | 3.67 |
| 10-15 | [7.8, 8.2) | 1715 | **1.74** | 2.85 | 2.94 | 2.98 | **2.02** | 3.37 | 3.46 | 3.51 |
| 15-20 | [8.2, 8.5) | 2702 | **1.62** | 2.48 | 2.57 | 2.61 | **1.93** | 3.00 | 3.09 | 3.15 |
| 20-25 | [8.5, 8.8) | 1369 | **1.36** | 2.27 | 2.36 | 2.41 | **1.77** | 2.87 | 2.96 | 3.01 |
| 25-30 | [8.8, 9.2) | 3175 | **1.20** | 2.13 | 2.22 | 2.27 | **1.59** | 2.77 | 2.87 | 2.93 |
| 30-35 | [9.2, 9.4) | 2472 | **1.05** | 1.81 | 1.90 | 1.95 | **1.46** | 2.49 | 2.59 | 2.65 |
| 35-40 | [9.4, 9.8) | 1239 | **0.98** | 1.79 | 1.86 | 1.90 | **1.37** | 2.44 | 2.53 | 2.57 |
| 40-45 | [9.8, 10.2) | 3007 | **0.97** | 1.64 | 1.72 | 1.76 | **1.33** | 2.30 | 2.39 | 2.46 |
| 45-50 | [10.2, 10.4) | 2378 | **0.96** | 1.55 | 1.61 | 1.65 | **1.26** | 2.13 | 2.23 | 2.30 |
| 50-55 | [10.4, 10.8) | 1083 | **1.07** | 1.63 | 1.67 | 1.69 | **1.34** | 2.15 | 2.22 | 2.26 |
| 55-60 | [10.8, 11.3) | 2840 | **1.09** | 1.58 | 1.61 | 1.64 | **1.33** | 2.05 | 2.13 | 2.18 |
| 60-65 | [11.3, 11.8) | 2584 | **1.18** | 1.61 | 1.63 | 1.66 | **1.42** | 2.01 | 2.07 | 2.12 |
| 65-70 | [11.8, 12.4) | 2568 | **1.31** | 1.65 | 1.65 | 1.67 | **1.58** | 1.98 | 2.02 | 2.05 |
| 70-75 | [12.4, 12.9) | 1971 | **1.50** | 1.80 | 1.81 | 1.83 | **1.80** | 2.14 | 2.16 | 2.20 |
| 75-80 | [12.9, 13.4) | 1797 | **1.69** | 1.93 | 1.93 | 1.94 | **2.01** | 2.28 | 2.29 | 2.31 |
| 80-85 | [13.4, 14.3) | 2966 | **1.98** | 2.14 | 2.12 | 2.12 | **2.33** | 2.53 | 2.50 | 2.51 |
| 85-90 | [14.3, 15.2) | 2225 | **2.51** | 2.62 | 2.56 | 2.55 | **2.89** | 3.09 | 3.04 | 3.03 |
| 90-95 | [15.2, 16.6) | 2267 | 3.23 | 3.30 | 3.20 | **3.17** | **3.59** | 3.82 | 3.74 | 3.71 |
| 95-100 | [16.6, 31.4] | 2270 | **5.24** | 5.55 | 5.40 | 5.33 | **5.66** | 6.14 | 6.02 | 5.96 |

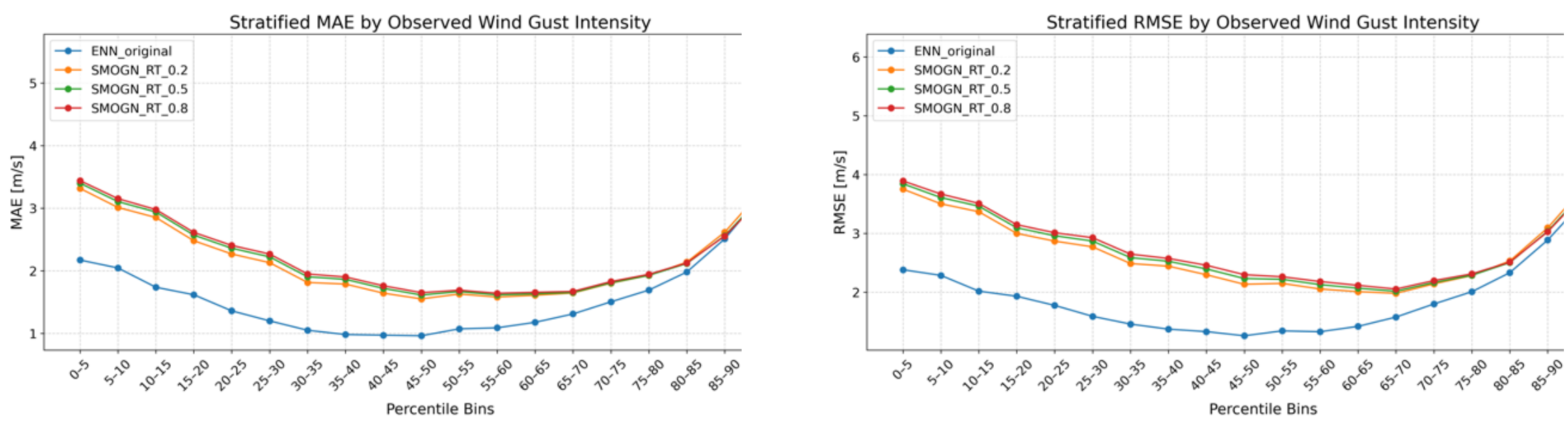

**Figure S4.** MAE and RMSE stratified by observed wind gust percentiles computed on test data. ENN_original refers to training without SMOGN. SMOGN_RT_0.2, SMOGN_RT_0.5 and SMOGN_RT_0.8 refer to training with SMOGN with RT values of 0.2, 0.5 and 0.8 respectively.

Table S3 and Fig. S4 show that while MAE after SMOGN was slightly lower than ENN_original at 90-95 percentile bin, ENN_original had the lowest MAE values in all other percentile bins. Also, RMSE did not show any improvement at any percentile with SMOGN. Finally, we attempted SMOGN for both over-sampling of minority values and under-sampling of majority values with an RT of 0.8. Since the pervious analyses showed lower prediction errors at higher gust percentiles for RT=0.8 than 0.2 and 0.5, we used RT of 0.8 to examine if simultaneous over and under-sampling performed better than oversampling only.

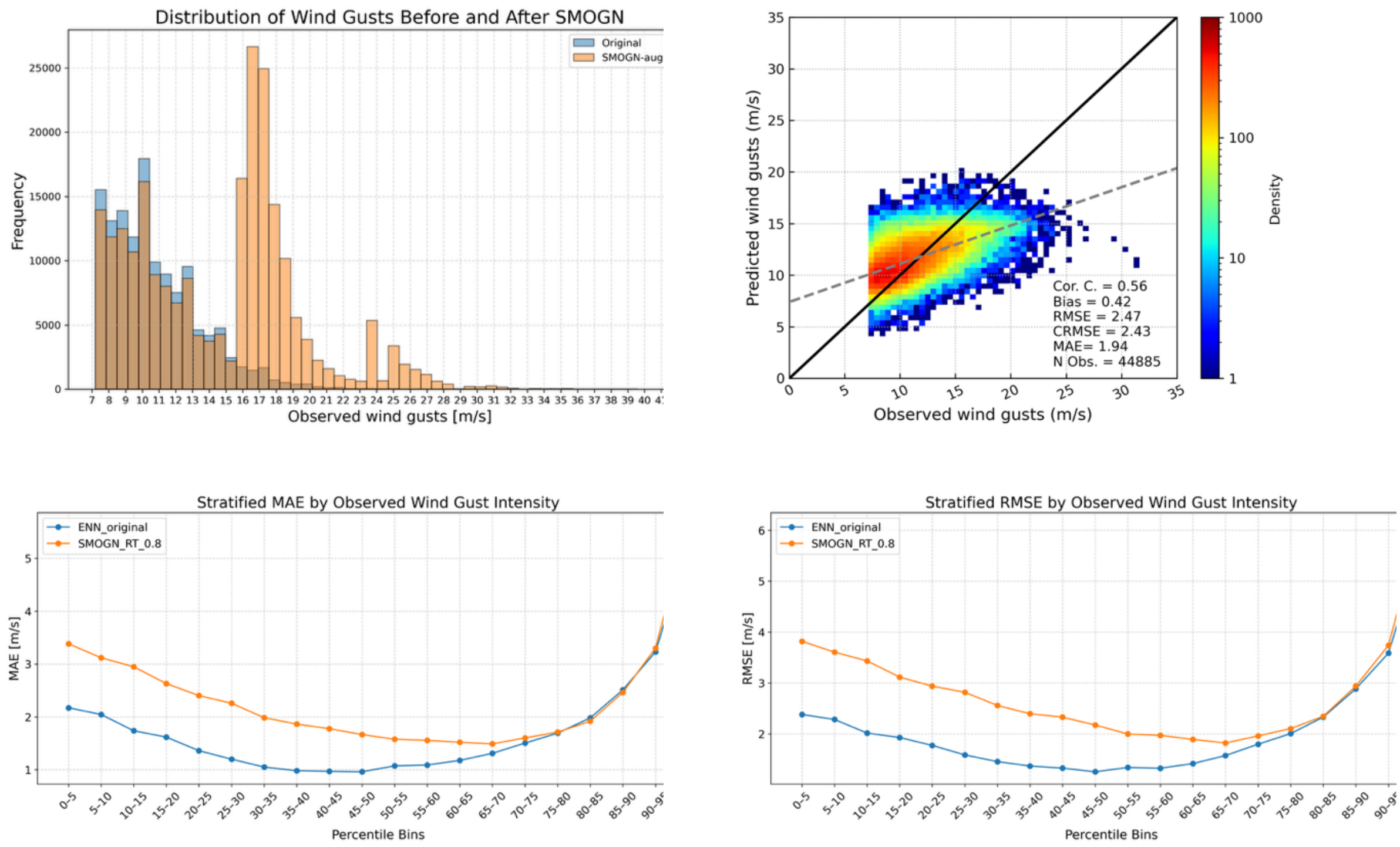

**Figure. S5.** (top left) Distribution of observed wind gusts in the training data before (original) and after data augmentation (SMOGN-augmented) for simultaneous over and under-sampling, (top right) wind gust predictions from ENN with SMOGN against observations, (bottom left) MAE and (bottom right) RMSE stratified by observed wind gust percentiles. ENN_original refers to training without SMOGN and SMOGN_RT_0.8 refers to training with SMOGN for both over and under-sampling with an RT value of 0.8.

Fig. S5 shows that over and under-sampling with SMOGN did not surpass predictions from ENN as well.

## 5. Miscellaneous

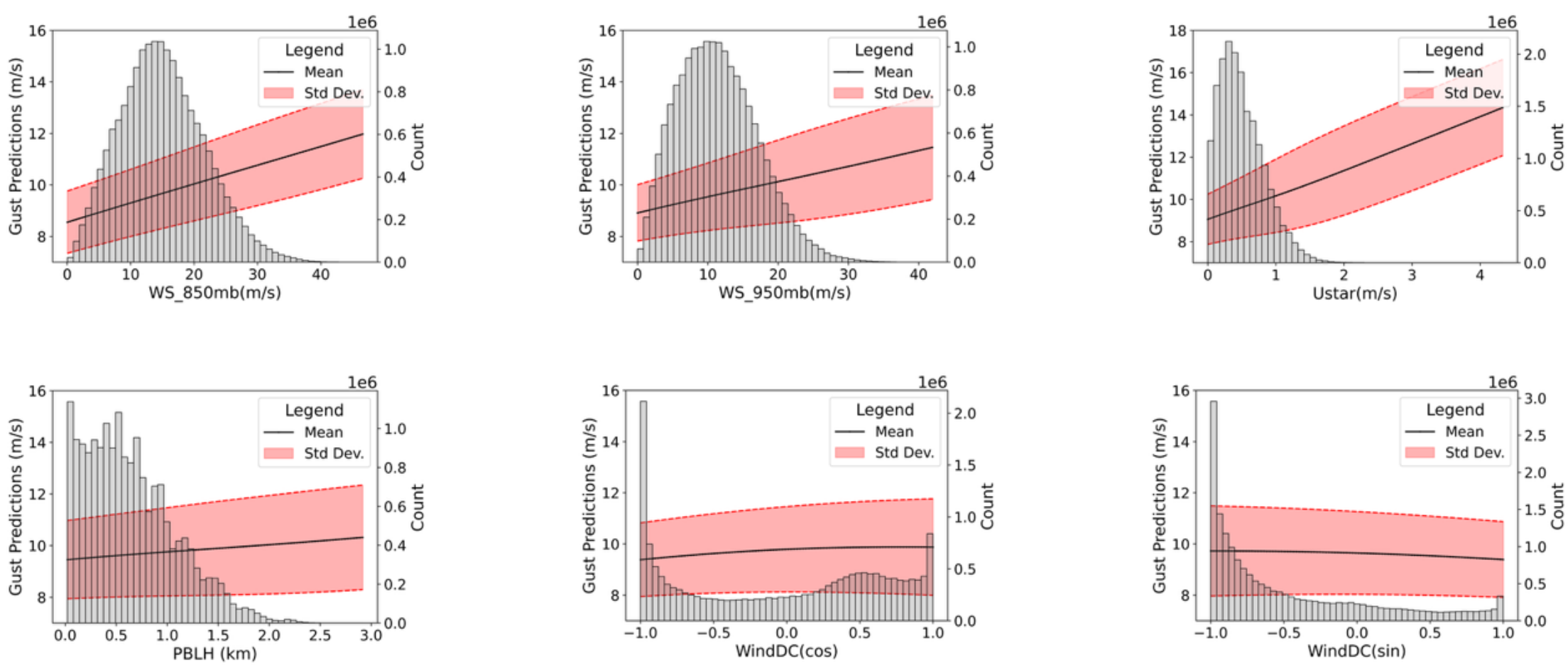

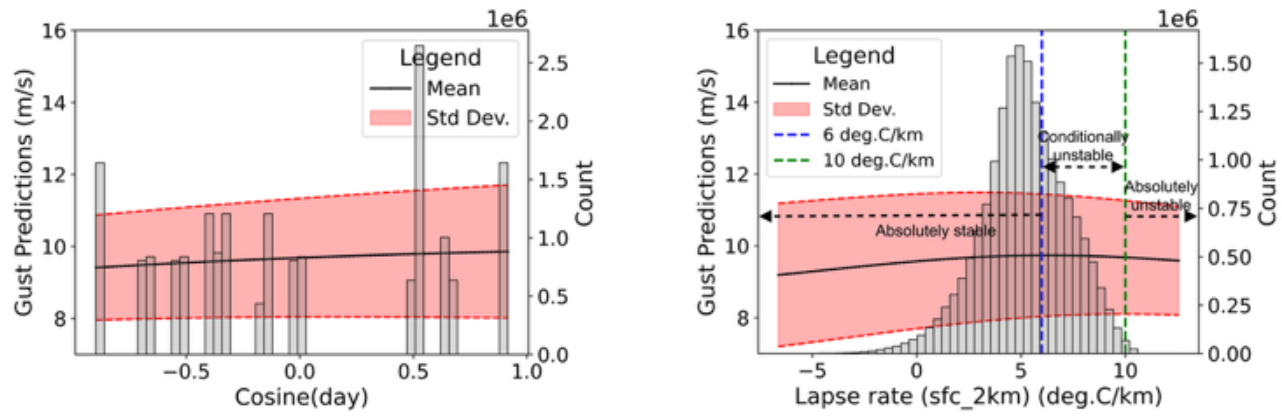

**Figure S6.** PDPs for predicted wind gusts for the following features: wind speed at 850 mb, wind speed at 950 mb, friction velocity (Ustar), planetary boundary layer height (PBLH), cosine of wind direction, sine of wind direction, cosine of day, lapse rate (surface to 2 km height).

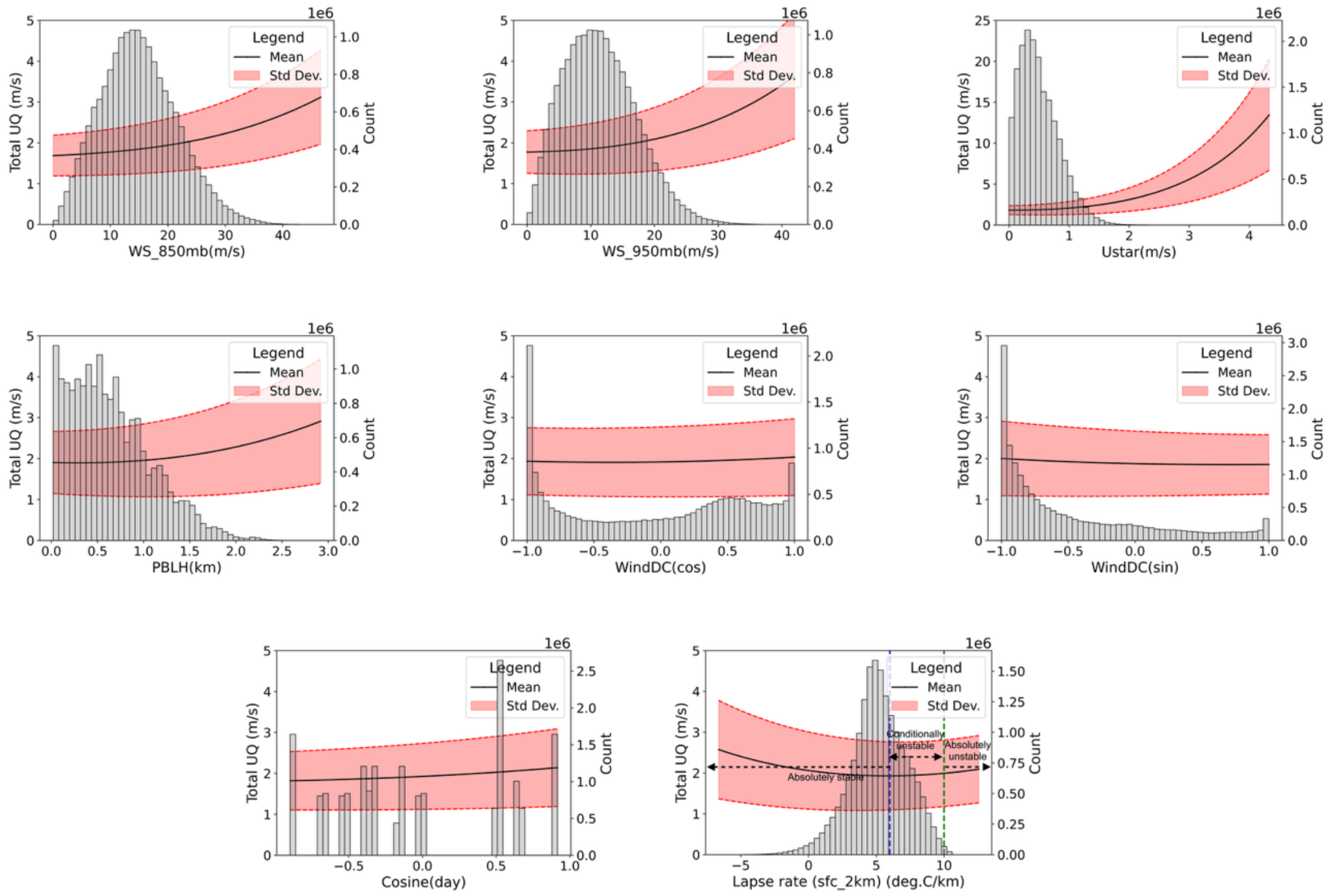

**Figure S7.** PDPs for total uncertainty for the following features: wind speed at 850 mb, wind speed at 950 mb, friction velocity (Ustar), planetary boundary layer height (PBLH), cosine of wind direction, sine of wind direction, cosine of day, lapse rate (surface to 2 km height).

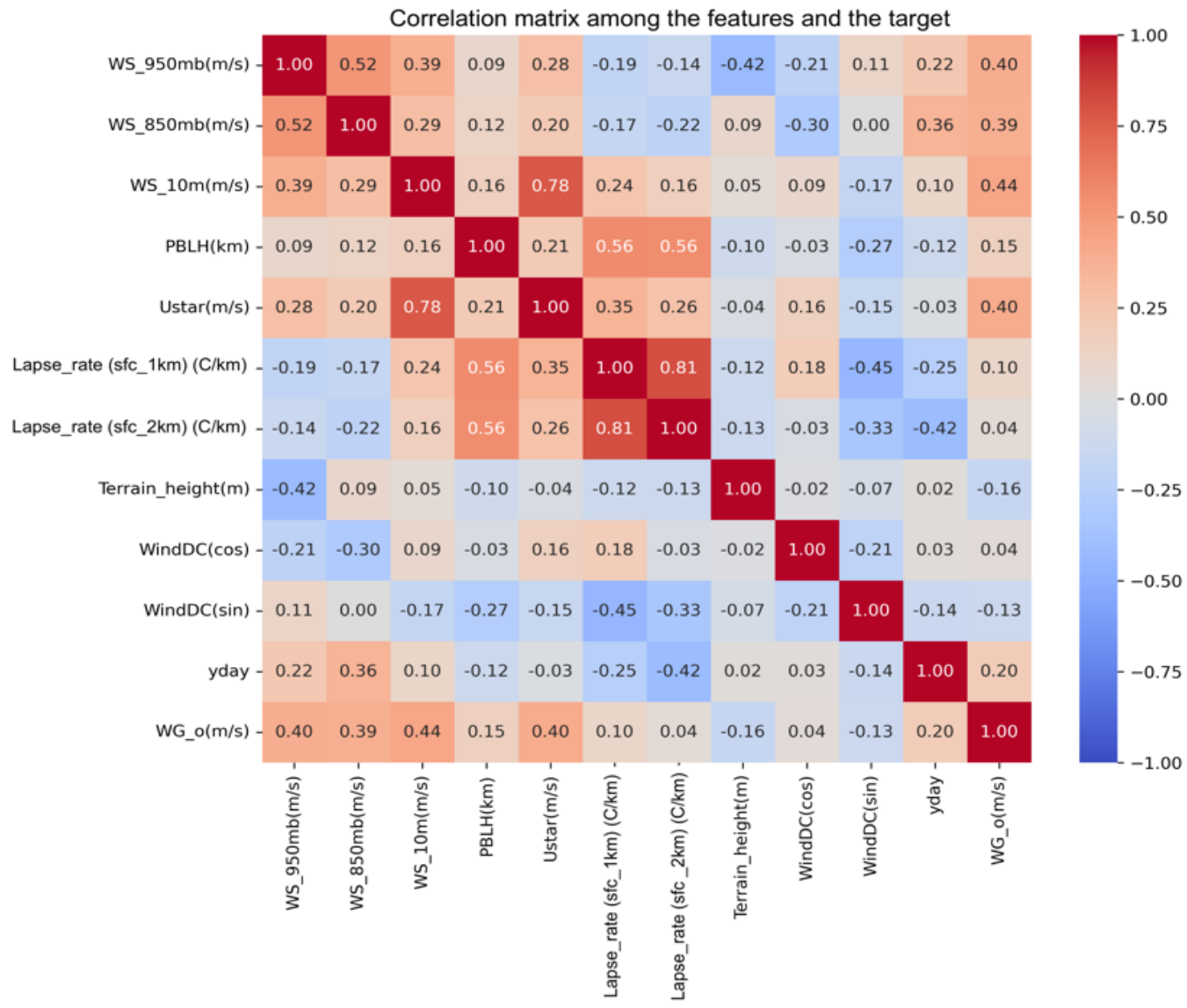

**Figure S8.** Correlation matrix among the input features and the target wind gust (WG_o).

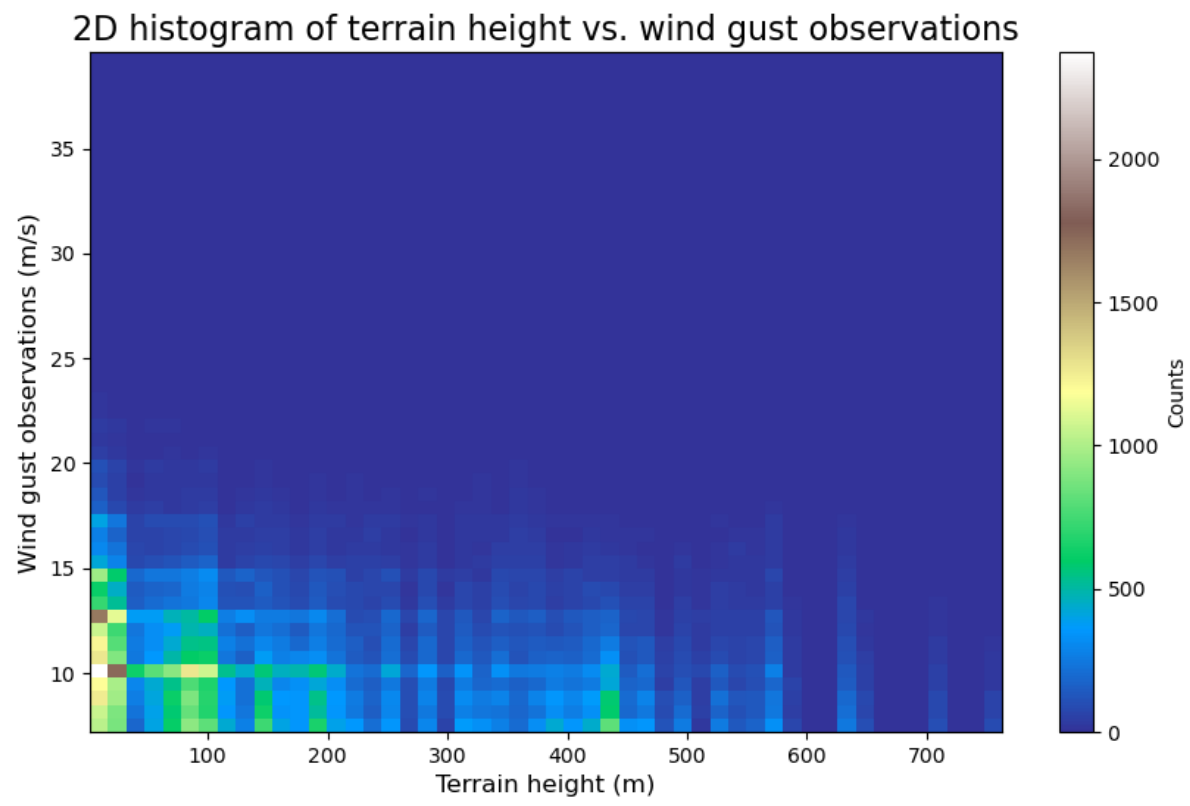

**Figure S9**. 2D histogram of wind gust observations and terrain height in the training data set.

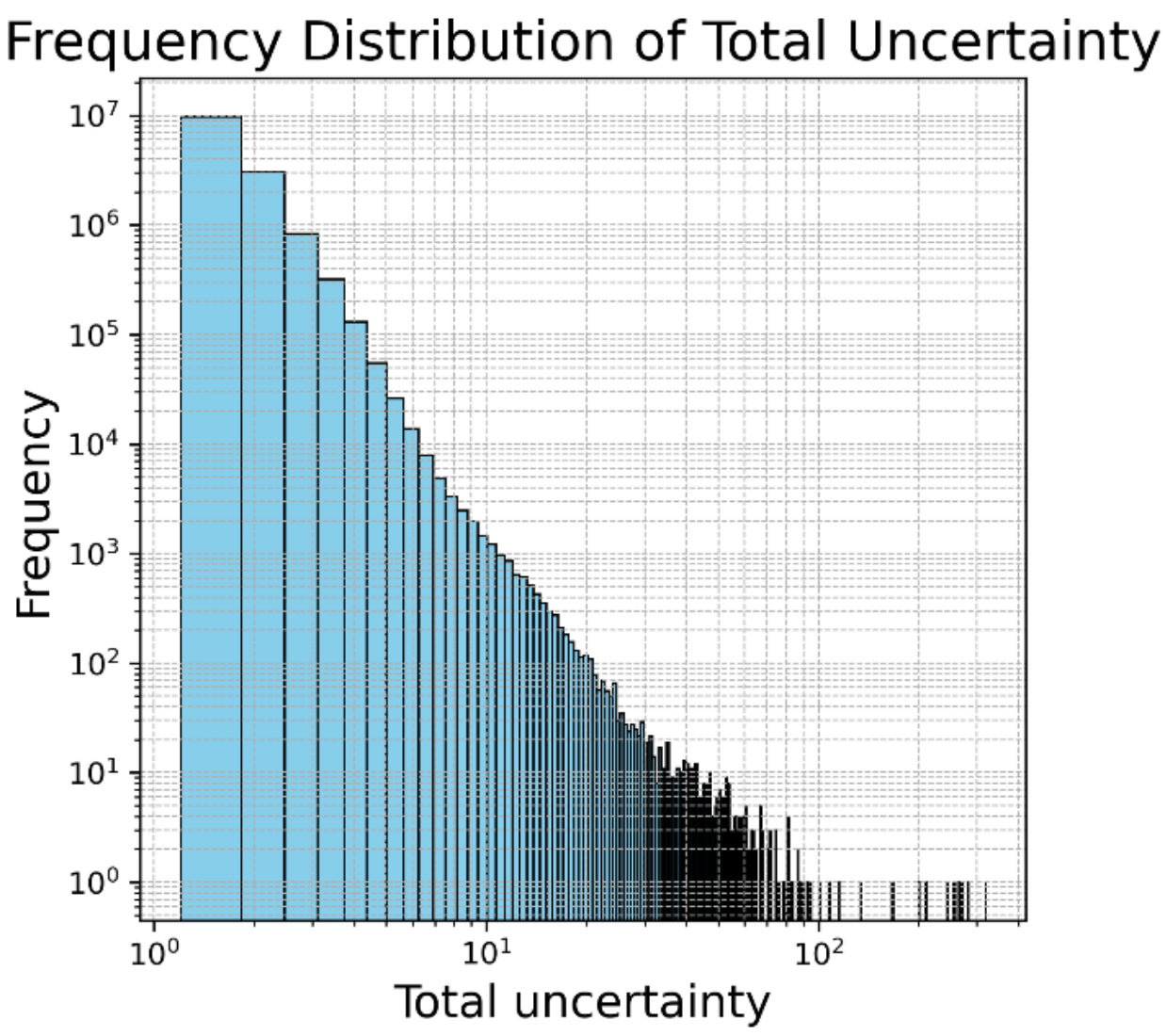

**Figure S10**. Distribution of total uncertainty estimated by ENN for the 11 test storms.